\documentclass[journal=sensors,article,accept,pdftex,moreauthors]{Definitions/mdpi}

\firstpage{1} 
\pubvolume{1}
\issuenum{1}
\articlenumber{0}
\pubyear{2024}
\copyrightyear{2024}
\datereceived{ } 
\daterevised{ } 
\dateaccepted{ } 
\datepublished{ } 
\hreflink{https://doi.org/} 

\usepackage{amsmath}
\usepackage{array} 
\usepackage{graphicx} 
\usepackage{makecell} 
\usepackage{multirow}
\usepackage{pdflscape} 
\usepackage{rotating}  
\usepackage{tikz} 
\usepackage{subcaption}
\usepackage{tabularx}
\usepackage{xcolor} 
\definecolor{amber}{rgb}{1.0, 0.75, 0.0}
\definecolor{darkgreen}{rgb}{0.0, 0.5, 0.0}
\definecolor{crimson}{rgb}{0.86, 0.08, 0.24}
\definecolor{navyblue}{rgb}{0.0, 0.0, 0.5}
\definecolor{teal}{rgb}{0.0, 0.5, 0.5}
\definecolor{customblue}{HTML}{4a86e8}
\definecolor{customgreen}{HTML}{34a853}
\definecolor{customorange}{HTML}{ff9900}
\definecolor{customred}{HTML}{ff0000}

\Title{InCrowd-VI: A Realistic Visual-Inertial Dataset for Evaluating SLAM in Indoor Pedestrian-Rich Spaces for Human Navigation}

\TitleCitation{InCrowd-VI: A Realistic Visual-Inertial Dataset for Evaluating SLAM in Indoor Pedestrian-Rich Spaces for Human Navigation}

\Author{Marziyeh Bamdad $^{1,2}$*\orcidA{}, Hans-Peter Hutter $^{1}$ and Alireza Darvishy $^{1}$}

\AuthorNames{Marziyeh Bamdad, Hans-Peter Hutter and Alireza Darvishy}

\AuthorCitation{Bamdad, M.; Hutter, HP.; Darvishy, A.}

\address{%
$^{1}$ \quad Institute of Applied Information Technology, Zurich University of Applied Sciences; marziyeh.bamdad@uzh.ch\\
$^{2}$ \quad Department of Informatics, University of Zurich;}

\corres{Correspondence: marziyeh.bamdad@uzh.ch}

\abstract{Simultaneous localization and mapping (SLAM) techniques can be used to navigate the visually impaired, but the development of robust SLAM solutions for crowded spaces is limited by the lack of realistic datasets. To address this, we introduce InCrowd-VI, a novel visual-inertial dataset specifically designed for human navigation in indoor pedestrian-rich environments. Recorded using Meta Aria Project glasses, it captures realistic scenarios without environmental control. InCrowd-VI features 58 sequences totaling a 5 km trajectory length and 1.5 hours of recording time, including RGB, stereo images, and IMU measurements. The dataset captures important challenges such as pedestrian occlusions, varying crowd densities, complex layouts, and lighting changes. Ground-truth trajectories, accurate to approximately 2 cm, are provided in the dataset, originating from the Meta Aria project machine perception SLAM service. In addition, a semi-dense 3D point cloud of scenes is provided for each sequence. The evaluation of state-of-the-art visual odometry (VO) and SLAM algorithms on InCrowd-VI revealed severe performance limitations in these realistic scenarios. Under challenging conditions, systems exceeded the required localization accuracy of 0.5 meters and the 1\% drift threshold, with classical methods showing drift up to 5-10\%. While deep learning-based approaches maintained high pose estimation coverage (>90\%), they failed to achieve real-time processing speeds necessary for walking pace navigation. These results demonstrate the need and value of a new dataset to advance SLAM research for visually impaired navigation in complex indoor environments. The dataset and associated tools are publicly available at https://incrowd-vi.cloudlab.zhaw.ch/.}

\keyword{visual SLAM; blind and visually impaired navigation; crowded indoor environments; dataset} 

\begin{document}

\section{Introduction}
\label{sec:intro}
Navigation in crowded indoor public spaces presents major challenges for blind and visually impaired (BVI) individuals. Systems that support this navigation require real-time user localization, detailed environmental mapping, and enhanced spatial awareness. Robust solutions are necessary to cope with unfamiliar settings and provide safe and more independent mobility for people with visual disabilities.  Simultaneous localization and mapping (SLAM) \cite{durrant2006simultaneous} offers promising capabilities for addressing these requirements. However, several hurdles must be overcome in order to make SLAM viable for visually impaired navigation, particularly in crowded public spaces. These settings are characterized by unpredictable pedestrian movements, varying lighting conditions, and reflective and transparent surfaces. Such dynamic and complex environments complicate reliable navigation significantly.

Although existing SLAM research has made significant advancements in domains such as robotics \cite{cadena2016past, pelleritodeep}, autonomous driving \cite{bresson2017simultaneous}, and aerial vehicles \cite{hanover2024autonomous, alberico2024structure}, these approaches do not adequately address the specific challenges of pedestrian-rich indoor navigation for the visually impaired. 

The lack of realistic datasets tailored for human navigation in crowded environments has been a significant barrier in the development of robust SLAM systems tailored for visually impaired navigation. Current datasets have several key limitations:
\begin{itemize}
    \item Most existing datasets focus on vehicle-based scenarios or controlled environments with limited dynamic elements.
    \item Available pedestrian-focused datasets like ADVIO and BPOD lack the combination of high crowd density and accurate ground truth needed for robust SLAM evaluation.
    \item Datasets that do include crowded scenes, such as NAVER LABS, are typically collected from fixed or robotic platforms that don't capture natural human motion patterns.
    \item Existing datasets do not comprehensively address the challenges of indoor navigation, such as varying crowd densities, motion transitions, and complex architectural features.
\end{itemize}
To address these gaps, we introduce InCrowd-VI, a visual-inertial dataset specifically designed for SLAM research for human navigation in human-crowded indoor environments. Unlike existing datasets, InCrowd-VI captures sequences recorded at diverse indoor public locations, such as airports, train stations, museums, university labs, shopping malls, and libraries, representing realistic human motion patterns at typical walking speeds. The recorded sequences feature diverse settings, including varying crowd densities (from pedestrian-rich to static environments) and complex architectural layouts such as wide-open spaces, narrow corridors, (moving) ramps, staircases, and escalators. They present various challenges characteristic of real-world indoor spaces, including frequent occlusions by pedestrians, variations in lighting conditions, and presence of highly reflective surfaces. The dataset was collected with Meta Aria Project glasses worn by a walking person in pedestrian-rich environments, and thus incorporates realistic human motion, behavior, and interaction patterns. 

The dataset has a total trajectory length of 4998.17 meters, with a total recording time of 1 h, 26 min, and 37 s. This dataset provides RGB images, stereo images, and IMU measurements. In addition, it includes a semi-dense 3D point cloud of scenes for further analysis. 
The ground-truth trajectories were provided by the Meta Aria project machine perception SLAM service \cite{somasundaram2023project}, which offers a reliable benchmark for evaluating the accuracy of the SLAM algorithms. 

 To demonstrate the value of InCrowd-VI, several state-of-the-art classical and deep learning-based approaches for visual odometry (VO) and SLAM systems were evaluated. The analysis revealed the severe performance degradation of these systems in crowded scenarios, large-scale environments, and challenging light conditions, highlighting the key challenges and opportunities for future research to develop more robust SLAM solutions for visually impaired navigation.

The contributions of this paper are as follows.
\begin{itemize}
    \item Introduction of InCrowd-VI, a novel visual-inertial dataset specifically designed for human navigation in indoor pedestrian-rich environments, filling a critical gap in existing research resources.
    \item Provision of ground-truth data, including accurate trajectories (approximately 2 cm accuracy) and semi-dense 3D point clouds for each sequence, enabling rigorous evaluation of SLAM algorithms.
    \item Evaluation of state-of-the-art visual odometry and SLAM algorithms using InCrowd-VI, revealing their limitations in realistic crowded scenarios.
    \item Identification of crucial areas for improvement in SLAM systems designed for visually impaired navigation in complex indoor environments.
\end{itemize}
\section{Related Work}
\label{sec:related_work}
Evaluation of visual SLAM systems requires comprehensive datasets that capture the complexity and variability of real-world environments. The existing benchmark datasets for SLAM can be categorized according to their operational domains. Depending on the domain, different sensory data and varying degrees of ground truth accuracy have been provided \cite{zhang2022hilti}. Various datasets have been proposed for different operational domains, each with distinct sensor platforms, settings, and challenges.
This section reviews state-of-the-art datasets, highlighting their characteristics and limitations in comparison with the newly proposed InCrowd-VI dataset. Datasets from robotics and autonomous systems as well as those focused on pedestrian odometry, are examined to assess their applicability to BVI navigation challenges. Table~\ref{tab:related_work} provides an overview of these datasets, summarizing their key features and comparing them to InCrowd-VI.
\begin{table}[H]
\caption{Comparison of representative datasets.\label{tab:related_work}}
	\begin{adjustwidth}{-\extralength}{0cm}
		\newcolumntype{L}{>{\arraybackslash}X}
		\begin{tabularx}{\fulllength}{llLLlLc}
			\toprule
			\textbf{Dataset} & \textbf{Environment} & \textbf{Carrier} & \textbf{Sensors} & \textbf{Crowd Density} & \textbf{Ground Truth} & \textbf{\# Sequence} \\
            \midrule
            \multicolumn{7}{c}{Robotics and autonomous systems}\\
			\midrule
            KITTI \cite{geiger2012we} & Outdoor & Car & High resolution cameras, LiDAR, GPS/IMU & Low & GPS/IMU & 22 \\
            EuRoC MAV \cite{burri2016euroc} & Indoor & Drone & Stereo cameras, IMU & Static & Motion capture & 11 \\ 
            PennCOSYVIO \cite{pfrommer2017penncosyvio} & In/Outdoor & Hand-held & Stereo cameras, IMU & Low & Fiducial markers & 4 \\
            Zurich Urban \cite{majdik2017zurich} & Outdoor & Quadrotor & High-resolution camera, GPS, and IMU & Not mentioned & Photogrammetric 3D reconstruction & 2 km \\
            InteriorNet \cite{li2018interiornet} & Indoor & Simulated cameras & Synthetic images, IMU & None & Synthetic & 15K \\
            TUM VI \cite{schubert2018tum} & In/Outdoor & Handheld & Stereo cameras, IMU & Low & Partial motion capture & 28 \\
            UZH-FPV \cite{delmerico2019we} & In/Outdoor & Quadrotor & Event and RGB cameras, IMU & None & Leica Nova MS60 TotalStation & 27+ \\
            Newer College \cite{ramezani2020newer} & Outdoor & Hand-held & Stereoscopic-inertial camera, LiDAR & None & 6DOF ICP localization & 4 \\
            VIODE \cite{minoda2021viode} & In/Outdoor & Simulated quadrotor UAV & Synthetic RGB cameras, IMU & High & Synthetic & 12 \\
            NAVER LABS \cite{lee2021large} & Indoor & A dedicated mapping platform & Cameras, laser scanners & Medium & LiDAR SLAM \& SFM & 5 datasets \\
            ConsInv \cite{bojko2022self} & In/Outdoor & Not mentioned & Monocular and stereo camera & Low & ORB-SLAM2 & 159 \\
            Hilti-Oxford \cite{zhang2022hilti} & In/Outdoor & Hand-held & Stereo cameras, IMU, LiDAR & Low & survey-grade scanner & 16 \\
            CID-SIMS \cite{zhang2023cid} & Indoor & Robot/Handheld & RGB-D camera, IMU, wheel odometry & Low & GeoSLAM & 22 \\
            \midrule
            \multicolumn{7}{c}{Pedestrian odometry dataset} \\
            \midrule
            Zena \cite{recchiuto2017dataset} & Outdoor & Head-mounted & 2D laser scanner, IMU & High & Step estimation process using IMU & a 35-min dataset \\
            ADVIO \cite{cortes2018advio} & In/Outdoor & Hand-held & Smartphone cameras, IMU & High & IMU-based + manual position fixes & 23 \\
            BPOD \cite{charatan2022benchmarking} & In/Outdoor & Head-mounted & Stereo cameras, laser distance meter & Not mentioned & Marker-based & 48 \\
            InCrowd-VI (ours) & Indoor & Head-worn & RGB and stereo cameras, IMU & High & Meta Aria Project SLAM service & 58 \\
			\bottomrule
		\end{tabularx}
	\end{adjustwidth}
\end{table}

\subsection{Robotics and autonomous systems} 
Datasets play a crucial role in advancing SLAM research in various domains. The KITTI dataset \cite{geiger2012we}, which is pivotal for autonomous vehicle research, has limited applicability to indoor pedestrian navigation, because it focuses on outdoor environments. Similarly, datasets like EuRoC MAV \cite{burri2016euroc}, TUM VI \cite{schubert2018tum}, and Zurich Urban \cite{majdik2017zurich}, although offering high-quality visual-inertial data, do not fully capture the challenges of indoor pedestrian navigation in crowded environments. The CID-SIMS dataset \cite{zhang2023cid}, recorded from a ground-wheeled robot, provides IMU and wheel odometer data with semantic annotations and an accurate ground truth. However, it lacks the complexity of pedestrian-rich environments. VIODE \cite{minoda2021viode} offers a synthetic dataset from simulated UAV navigation in dynamic environments; however, synthetic data cannot fully replace real-world data, particularly for safety-critical applications \cite{liu2021simultaneous,tosi2024nerfs}. The ConsInv dataset \cite{bojko2022self} evaluates the SLAM systems with controlled dynamic elements. However, its controlled nature fails to capture the complexity of human-crowded public spaces. Moreover, its ground-truth method using ORB-SLAM2 with masked dynamic objects does not accurately represent the challenges faced by SLAM systems in crowded real-world settings.

\subsection{Pedestrian Odometry Dataset}
The Zena dataset \cite{recchiuto2017dataset} provides laser scan data and IMU measurements from helmet and waist-mounted sensors to support research on human localization and mapping in outdoor scenarios. However, the lack of visual data limits its utility in the assessment of visual SLAM systems. The Brown Pedestrian Odometry Dataset (BPOD) \cite{charatan2022benchmarking} provides real-world data from head-mounted stereo cameras in diverse indoor and outdoor environments. Although it captures challenges, such as rapid head movement and image blur, BPOD does not specifically focus on crowded indoor environments. ADVIO \cite{cortes2018advio} is a dataset for benchmarking visual-inertial odometry using smartphone sensors in various indoor and outdoor paths. Although valuable for general visual-inertial odometry, the use of handheld devices limits its ability to capture natural head movements and gait patterns, which are crucial for realistic navigation scenarios. Additionally, ADVIO's ground truth, which is based on inertial navigation with manual position fixes, may have accuracy limitations.

Although these datasets provide valuable resources for evaluating SLAM systems, they still need to fully address the specific challenges for human navigation in indoor pedestrian-rich environments. InCrowd-VI fills critical gaps by providing the following:
\begin{itemize}
    \item Natural human motion patterns captured through head-worn sensors represent visually impaired navigation scenarios better than the data collected using robotic or handheld devices in existing datasets.
    \item Comprehensive coverage of crowd densities ranging from empty to heavily crowded, allowing systematic evaluation of SLAM performance under varying levels of dynamic obstacles.
    \item Realistic indoor scenarios, including challenging architectural features and environmental conditions that are critical for practical navigation applications.
    \item High-quality ground truth trajectories even in crowded scenes, enabling precise evaluation of SLAM performance in challenging dynamic environments.
    \item Long-duration sequences that test system stability and drift in real-world navigation scenarios.
\end{itemize}
\section{InCrowd-VI Dataset}
\label{sec:incrowd-vi_dataset}
The InCrowd-VI dataset comprises 58 sequences with a total trajectory length of 4,998.17 meters and a total recording time of 1 hour, 26 minutes, and 37 seconds. The dataset consists of approximately 1T GB of extracted data and approximately 100 GB of raw data across these sequences. Sequences were captured across diverse indoor environments, including airports, train stations, museums, university buildings, shopping malls, and university labs and libraries. The dataset is categorized by crowd density (High: >10 pedestrians per frame, Medium: 4-10 pedestrians, Low: 1-3 pedestrians, and None: no pedestrians). It includes environmental challenges, such as lighting variations and reflective surfaces. Individual sequence lengths range from 12.20m to 348.84m, with recording durations spanning from 34 seconds to 5 minutes and 50 seconds. Additional detailed descriptions of each sequence are documented on the dataset website.

This section presents the InCrowd-VI dataset, specifically developed for evaluating SLAM in indoor pedestrian-rich environments for human navigation. The sensor framework used for data collection is first described, followed by an outline of the methodology employed to create the dataset. The process of obtaining and validating the ground-truth data is then explained. Finally, the captured sequences and the various challenges they represent are detailed.
%
%
\subsection{Sensor Framework}
The choice of platform for data collection was determined on the basis of the intended application of the SLAM system to be evaluated using the dataset. In the context of visually impaired navigation, wearable platforms are particularly appropriate because they can effectively capture human motion patterns during movements. Consequently, we employed a head-worn platform to collect the data. Head-mounted devices offer the advantage of capturing a forward-facing view, which is crucial for navigation, and more accurately representing how a visually impaired individual might scan their environment, including natural head movements and areas of focus. In this study, we utilized Meta Aria glasses as our sensor platform.

The Meta Aria glasses feature five cameras, including two mono scene cameras with less overlapping and large field of view, one RGB camera, and two eye-tracking (ET) cameras\footnote{https://facebookresearch.github.io/projectaria\_tools/docs/tech\_spec/hardware\_spec}. Additionally, the glasses are equipped with several non-visual sensors, including two Inertial Measurement Units (IMUs), a magnetometer, a barometer, a GPS receiver, and both Wi-Fi and Bluetooth beacons. The glasses also include a seven-channel spatial microphone array with a 48 kHz sampling rate, which can be configured to operate in stereo mode with two channels. It should be noted that the InCrowd-VI dataset includes data from only the RGB and mono cameras as well as IMU measurements. Other sensor data are not included in the dataset. One of the key features of Meta Aria glasses is their support for multiple recording profiles, allowing users to select which sensors to record with and configure their settings accordingly. This flexibility makes these glasses particularly suited for diverse experimental conditions and requirements. Table~\ref{tab:camera_specs} summarizes the specifications of the five cameras on Aria glasses.
\begin{table}
  \scriptsize
  \centering
  \begin{tabular}{@{}llllllc@{}}
    \toprule
    Camera & HFOV & VFOV & IFOV & Max resolution & FPS & Shutter \\
    \midrule
    Mono (x2) & 150 & 120 & 0.26 & 640x480 & 30 & global \\
    RGB (x1) & 110 & 110 & 0.038 & 2880x2880 & 30 & rolling \\
    ET (x2)	& 64 & 48 & 0.2 & 640x480 & 90 & global \\
    \bottomrule
  \end{tabular}
  \caption{Specifications of the cameras on Meta Aria glasses \cite{projectariasensorspec}, with the horizontal field of view (HFOV) in degrees, vertical field of view (VFOV) in degrees, instantaneous field of view (IFOV) in degrees per pixel, and maximum resolution in pixels. FPS represents the maximum frame rate. Note that InCrowd-VI only includes data from the RGB and Mono cameras.}
  \label{tab:camera_specs}
\end{table}
%
%
\subsubsection{Sensor Calibration}
The Meta Aria glasses used in our dataset underwent rigorous factory calibration for both intrinsic and extrinsic parameters. The factory calibration provides fundamental sensor parameters:
\paragraph{\textbf{Camera Intrinsics} \cite{CameraIntrinsicModelsforProjectAriadevices}}
The Meta Aria camera intrinsic model establishes a mapping between 3D world points in camera coordinates and their corresponding 2D pixels on the sensor. For the FisheyeRadTanThinPrism model, Meta Aria bases this mapping on polar coordinates of 3D world points. The mapping begins with a 3D world point in the device frame $P_d$, which transforms to the camera's local frame through:

\begin{equation}
P_c = (x, y, z) = T_{device}^{camera}P_d
\end{equation}

The resulting polar coordinates $\Phi = (\theta, \phi)$ satisfy:

\begin{equation}
x/z = \tan(\theta)\cos(\phi)
\end{equation}

\begin{equation}
y/z = \tan(\theta)\sin(\phi)
\end{equation}

These coordinates then map to a 2D pixel through the projection mapping:

\begin{equation}
p = f(\phi)
\end{equation}

with its inverse projection defined as:

\begin{equation}
\Phi = f^{-1}(p)
\end{equation}

The FisheyeRadTanThinPrism model used by Meta Aria extends beyond basic projection by incorporating thin-prism distortion ($tp$) on top of the Fisheye62 model. The complete projection function takes the form:

\begin{equation}
u = f_x \cdot (u_r + t_x(u_r,v_r) + tp_x(u_r,v_r)) + c_x
\end{equation}

\begin{equation}
v = f_y \cdot (v_r + t_y(u_r,v_r) + tp_y(u_r,v_r)) + c_y
\end{equation}

In these projection equations, $(u_r, v_r)$ represent the radially distorted coordinates, $(f_x, f_y)$ are focal lengths, and $(c_x, c_y)$ denote principal points. The terms $t_x, t_y$ account for tangential distortion, while $tp_x, tp_y$ represent the thin-prism distortion components.

The Meta Aria model defines the thin-prism distortion through:

\begin{equation}
tp_x(u_r,v_r) = s_0r(\theta)^2 + s_1r(\theta)^4
\end{equation}

\begin{equation}
tp_y(u_r,v_r) = s_2r(\theta)^2 + s_3r(\theta)^4
\end{equation}

where the coefficients $s_0, s_1, s_2, s_3$ serve as the thin-prism distortion parameters.

%
%
\paragraph{\textbf{IMU Calibration} \cite{metaariaimu}}
The Meta Aria project employs an affine model for IMU calibration, where the raw sensor readouts from both accelerometer and gyroscope are compensated to obtain the actual acceleration and angular velocity measurements. For the accelerometer measurements, the compensation is defined as:

\begin{equation}
a = M_a^{-1}(s_a - b_a)
\end{equation}

Similarly, for the gyroscope measurements:

\begin{equation}
\omega = M_g^{-1}(s_g - b_g)
\end{equation}

where $s_a$ and $s_g$ represent the raw sensor readouts from the accelerometer and gyroscope respectively, $M_a$ and $M_g$ are scale matrices assumed to be upper triangular to maintain no global rotation from the IMU body frame to the accelerometer frame, and $b_a$ and $b_g$ are the bias vectors.

The inverse of these compensation equations defines how the sensor readouts relate to the actual physical quantities:

\begin{equation}
s_a = M_aa + b_a
\end{equation}

\begin{equation}
s_g = M_g\omega + b_g
\end{equation}

The Meta Aria IMU system implements specific saturation limits to handle measurement bounds. These limits vary between the left and right IMUs, with the accelerometers bounded at 4g and 8g respectively, and the gyroscopes at 5000 and 1000 degrees per second respectively.
%
%
\paragraph{\textbf{Extrinsic Calibration}}
The Meta Aria glasses provide factory-calibrated extrinsic parameters that model the 6-DoF pose among the sensors. The extrinsic calibration expresses the relative poses between sensors using SE(3) Lie group representations \cite{3DCoordinateFrameConventionsforProjectAriaGlasses}, following the Hamilton quaternion convention. The device frame serves as the reference coordinate system and is aligned by default with a left stereo camera. For cameras, the local coordinate frame is defined based on the optical axis and the entrance pupil of the lens, with its origin at the center of the entrance pupil. For IMUs, the coordinate systems have their origins at the position of the accelerometer, oriented along the accelerometer's sensitive axis, and orthogonalized to compensate for the sensor orthogonality error. The Meta Aria project assumes that the accelerometer and gyroscope for each IMU are co-located; thus, they share the same extrinsic parameters, which simplifies the sensor fusion process. These calibration parameters remain fixed because they are determined during factory calibration and are provided as part of the calibration data of the device. This factory-provided calibration ensures reliable spatial relationships between sensors, which is crucial for accurate visual-inertial data collection and subsequent SLAM processing.
%
%
\subsection{Methodology}
The data collection process was carefully designed to capture the diverse challenges of real-world indoor navigation. The strength of the dataset lies in its comprehensive representation of these challenges, which include frequent pedestrian occlusions, varying crowd densities, complex architectural layouts, wide open spaces, narrow corridors, (moving) ramps, staircases, escalators, texture-poor scenes, lighting variations, and highly reflective surfaces. These environments range from densely populated areas with more than 10 pedestrians per frame to empty cluttered spaces, offering a wide variety of scenarios for evaluating SLAM systems.

During data collection, tactile paving (textured ground surfaces that BVI individuals can feel with their feet and cane) was followed where available, to mimic the walking patterns of visually impaired individuals and to enable later visual inspection of the accuracy of 3D point clouds, particularly in high-traffic areas. Data collection was conducted with the necessary permission from the relevant authorities, ensuring that ethical considerations were addressed.

To ensure the authenticity of the pedestrian movement patterns, we prioritized capturing organic and unscripted interactions during data collection. Rather than staging specific scenarios or providing explicit instructions to pedestrians, we adopted an observational approach that allowed for natural human behavior. Data was collected from real-world public spaces, including train stations, airports, shopping malls, and university campuses, during typical hours of pedestrian activity. Participants wearing Meta Aria glasses were instructed to navigate these spaces as they normally would, without any artificial constraints on their movement or interactions with the environment. This approach ensured that pedestrian movements were spontaneous and not artificially controlled, whereas the dataset captured genuine navigation challenges, including unpredictable pedestrian interactions. Walking speeds and movement patterns reflected real-world variability, with an average walking speed of 0.75 m/s further emphasizing the natural navigation patterns observed in crowded spaces. By allowing pedestrians to move freely and capture their interactions, we ensured that the dataset reflected the inherent complexity and dynamism of indoor human navigation.
%
%
\subsection{Ground-Truth}
The ground-truth trajectory and 3D reconstruction of the scene were produced using the Meta Aria machine perception SLAM service \cite{somasundaram2023project}. This service provides device trajectories generated by state-of-the-art VIO and SLAM systems, followed by offline postprocessing and refinement. It uses multiple sensors to improve accuracy and robustness, taking advantage of the precise knowledge of the sensor models, timing, and rigidity of Meta Aria devices. This allows for robust localization even under challenging real-world conditions such as fast motion, low or highly dynamic lighting, partial or temporary occlusion of cameras, and a wide range of static and dynamic environments. 

Although the Meta Aria machine perception SLAM service achieves high accuracy, it does so under conditions that are not feasible for real-time visually impaired navigation: it operates offline with server-side processing, utilizes the full sensor suite, and performs extensive post-processing. By contrast, practical navigation systems for the visually impaired must operate in real time on resource-constrained wearable devices, provide immediate reliable feedback, and maintain robustness without the benefit of post-processing or cloud computation.

The trajectories produced by the Meta SLAM service have a typical global RMSE translation error of no more than 1.5 cm in room-scale scenarios. Additionally, Meta SLAM service provides a semi-dense point cloud of the scene, accurately reconstructing the static portion of the environment, even in highly dynamic and challenging situations \cite{somasundaram2023project}. 

\begin{figure}[H]
  \centering
      \begin{subfigure}{0.32\linewidth}
        \includegraphics[width=\textwidth]{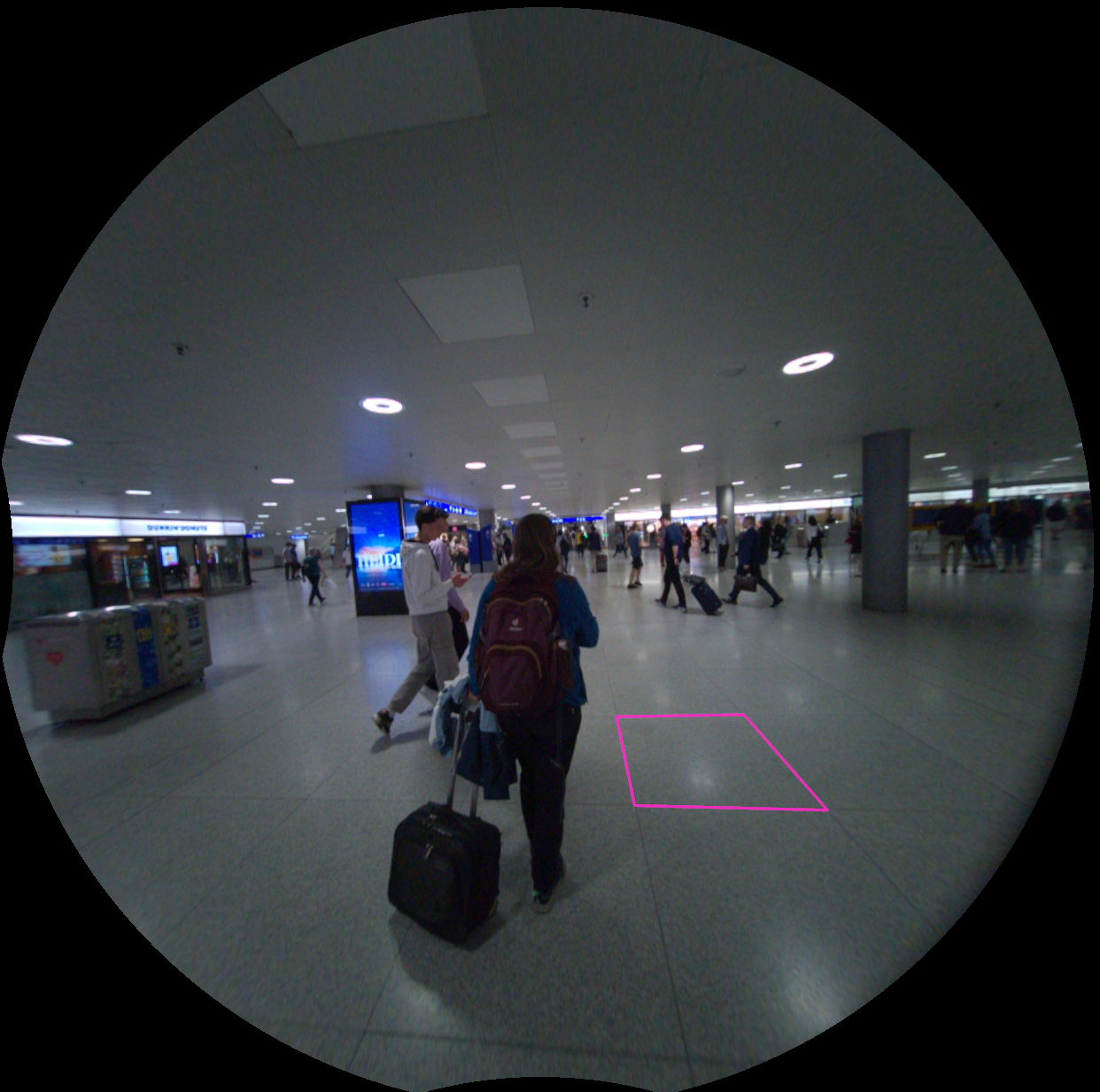}
        \label{fig:manual_measurement_scene}
      \end{subfigure}
      \hfill
      \begin{subfigure}{0.32\linewidth}
        \includegraphics[width=\textwidth]{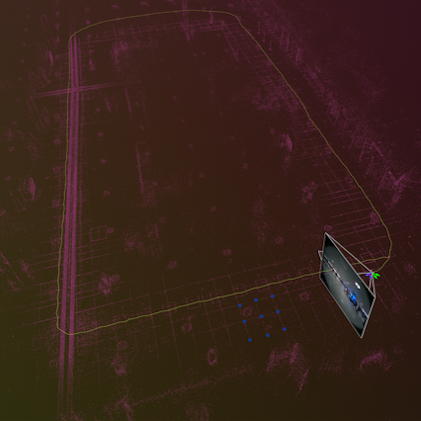}
        \label{fig:manual_measurement_map}
      \end{subfigure}
      \hfill
      \begin{subfigure}{0.32\linewidth}
        \includegraphics[width=\textwidth]{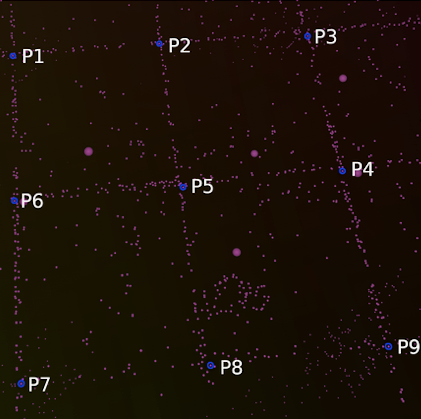}
        \label{fig:manual_measurement_zoom}
      \end{subfigure}
\caption{Sample of manual measurement process for ground-truth validation. Left: Real-world scene with a landmark floor tile highlighted by pink rectangle. Middle: Full 3D point cloud map of the scene with four adjacent floor tiles marked in blue. Right: Zoomed view of the marked corner of the tiles in the point cloud used for measurement.\label{fig:manual_measurement}}
\end{figure}

\begin{figure}[H]
\includegraphics[width=0.8\linewidth]{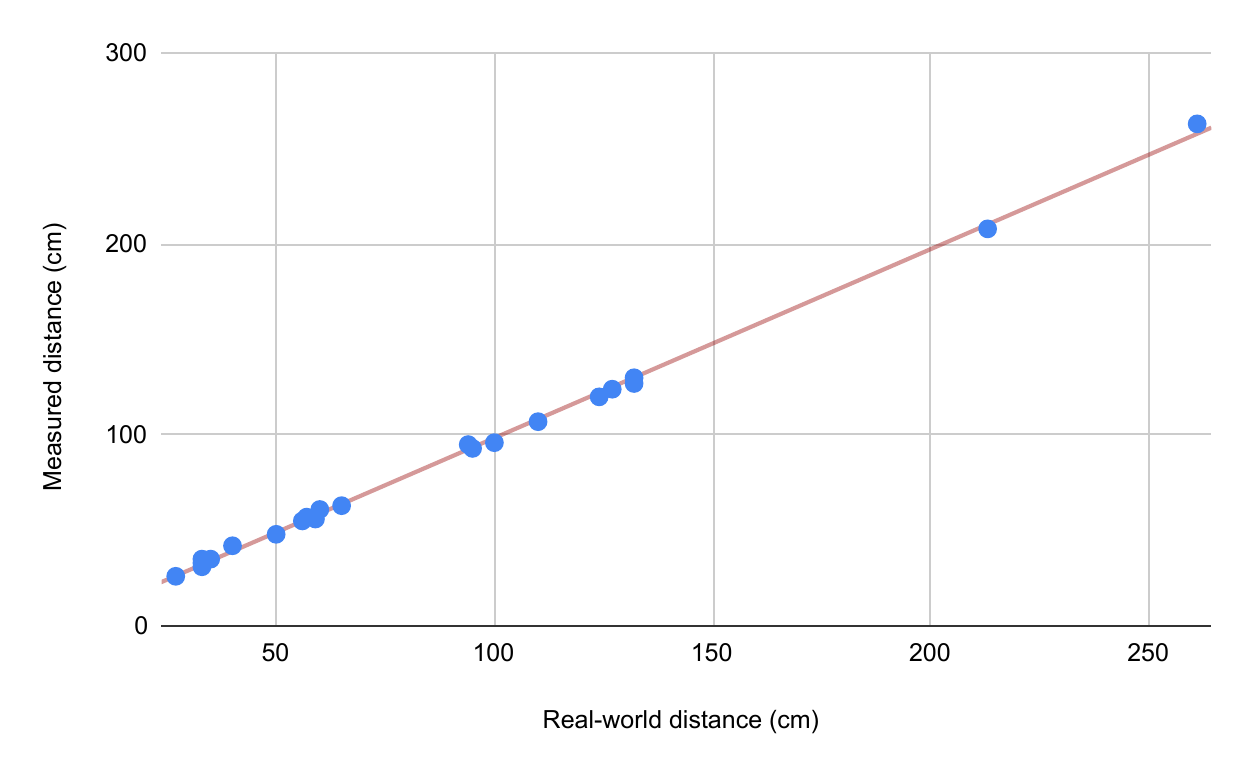}
\caption{ Correlation between real-world measurements and point-cloud-derived distances in challenging sequences, where state-of-the-art SLAM systems exhibited failure or suboptimal performance. The scatter plot demonstrates a strong linear relationship between real-world and measured distances (in centimeters), with an average error of 2.14 cm, standard deviation of 1.46 cm, and median error of 2.0 cm.\label{fig:real_vs_measured}}
\end{figure}

The accuracy of this ground truth was further validated through manual measurements in several challenging scenarios, with results indicating a mean absolute error of approximately 2 cm, aligning with the reported accuracy of the Meta SLAM service.
To validate the ground truth, a method was used that leverages the core principle of SLAM systems: the joint optimization of the 3D map and camera trajectory \cite{schops2019bad}. This approach involved identifying easily recognizable landmarks across the trajectory in both the real-world environment and semi-dense map. The 3D coordinates of these landmarks were recorded from the point cloud and the distances between them were calculated using the Euclidean distance formula. These calculated distances were then compared with actual measurements taken in the real world, allowing for a direct assessment of the accuracy of the map and reliability of the estimated trajectory. Figure~\ref{fig:manual_measurement} illustrates an example of the manual measurement process. 
Initially, manual measurements were conducted on selected crowded sequences. Following the evaluation of the state-of-the-art VO and SLAM systems presented in Section~\ref{sec:evaluation_results}, we conducted additional manual measurements, specifically focusing on sequences where all systems exhibited failure or suboptimal performance, according to the metrics defined in Section~\ref{sec:evaluation_metrics}. We used a variety of objects in each sequence, such as floor tiles, doors, advertising boards, and manhole covers, distributed across different spatial orientations throughout the trajectories to ensure robust validation. 
Figure~\ref{fig:real_vs_measured} presents the relationship between real-world and measured distances from the second round of manual measurements. The strong linear correlation (indicated by the red trend line) and tight clustering of points around this line demonstrates that the Meta SLAM service maintains its accuracy even in challenging scenarios where contemporary SLAM systems struggle. The plot, incorporating measurements ranging from 30 cm to over 250 cm, shows that the reconstruction accuracy remained stable regardless of the measured distance, with deviations typically within 2 cm of the expected values.
\unskip
   
\begin{figure}[H]
\includegraphics[width=0.9\linewidth]{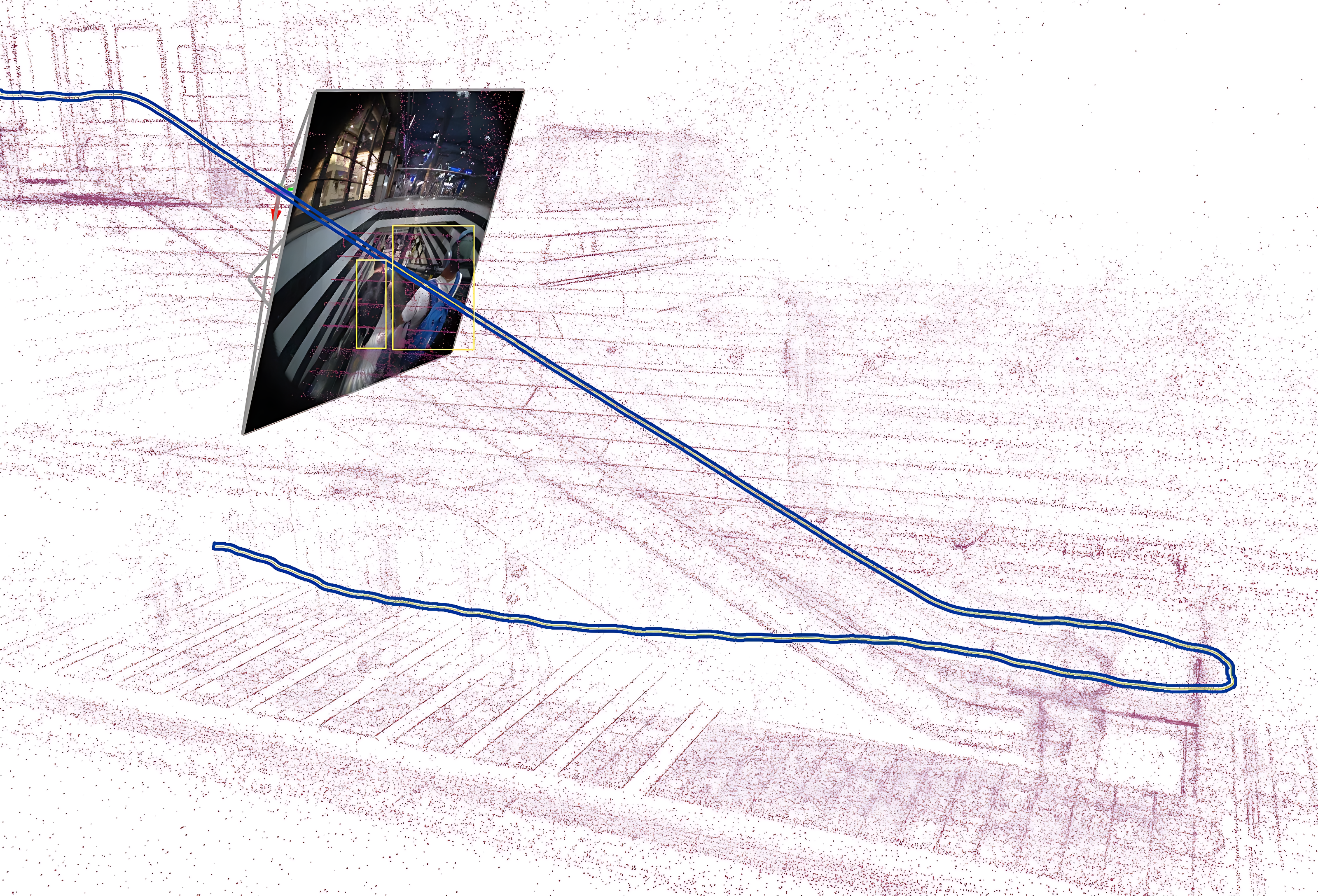}
\caption{Refined 3D reconstruction demonstrating the removal of dynamic pedestrians that initially appeared static relative to the camera on the escalator.\label{fig:removing_dynamic_objects}}
\end{figure}

It is important to note that the manual measurement process itself introduces some level of error owing to the challenges in precisely identifying corresponding points. Despite this, the observed errors, which are slightly higher than the reported typical error, remain within a reasonable range, suggesting that the Meta SLAM service performs adequately in this specific scenario. 
In addition to quantitative metrics, a qualitative visual inspection of the estimated trajectories and maps was performed. This involved assessing the alignment of landmarks, plausibility of trajectories, and removal of moving pedestrians in the scene. Figure~\ref{fig:removing_dynamic_objects} illustrates the capability of the Meta Aria machine perception SLAM service to handle dynamic objects by showcasing a scenario where a pedestrian initially appears static relative to the camera while on an escalator but subsequently becomes dynamic. The image presents the refined 3D reconstruction produced by the SLAM service, which successfully identifies and removes dynamic pedestrians from the final point cloud, leaving only static elements of the environment. The dynamic object removal process improves the accuracy of 3D map reconstruction and trajectory estimation by ensuring that moving and temporary stationary objects are not incorporated into the static environment representation or ground truth.
\unskip

%
%
\subsection{Sequences and Captured Challenges}
\begin{figure}[H]
\centering
\scalebox{0.47}{ 
    \begin{minipage}[c]{\textwidth}
        \centering
        \begin{minipage}[c]{0.33\textwidth}
            \centering
            \includegraphics[width=\textwidth, height=0.33\textheight,keepaspectratio]{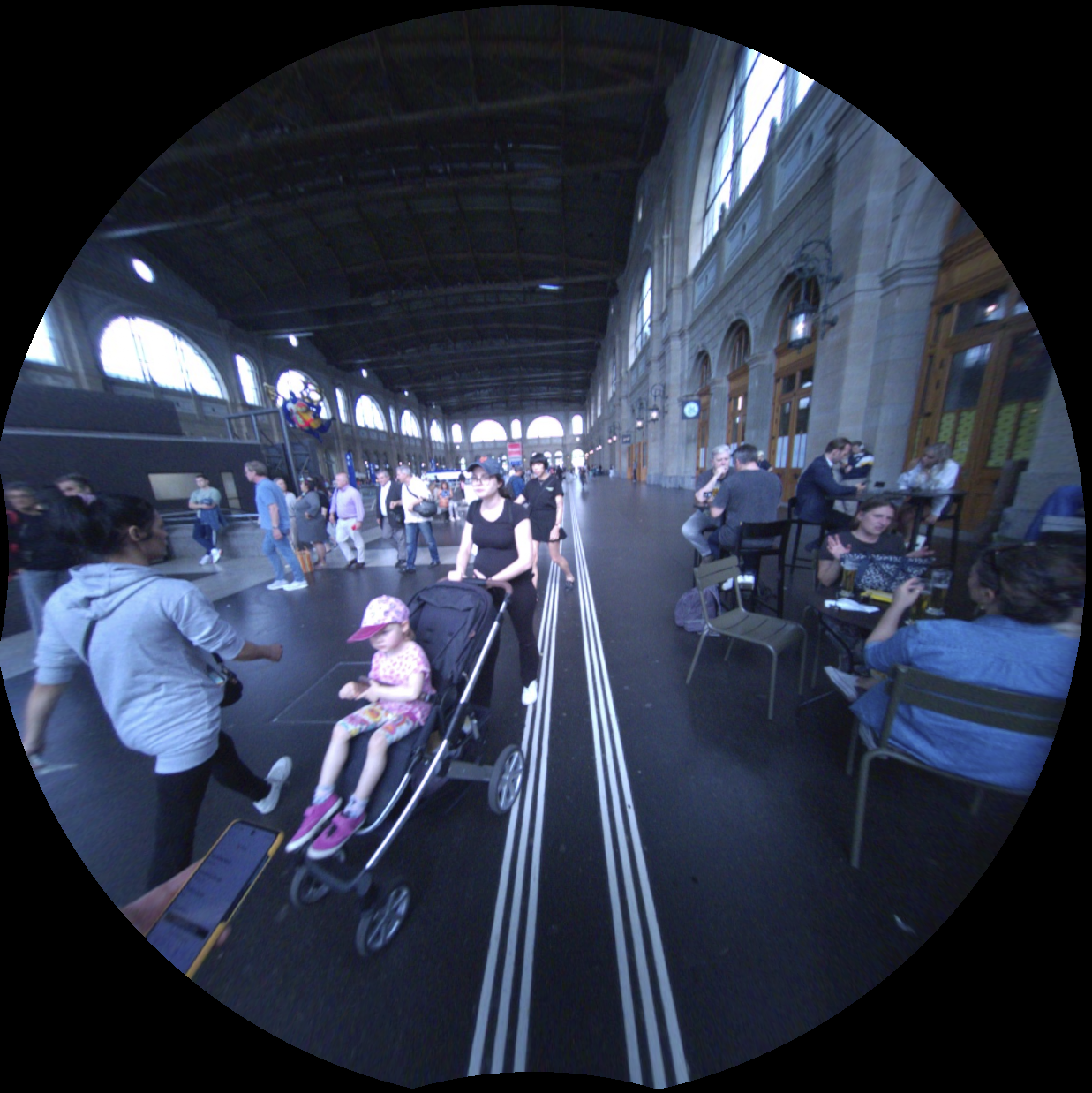}
        \end{minipage}\hfill
        \begin{minipage}[c]{0.33\textwidth}
            \centering
            \includegraphics[width=\textwidth, height=0.33\textheight,keepaspectratio]{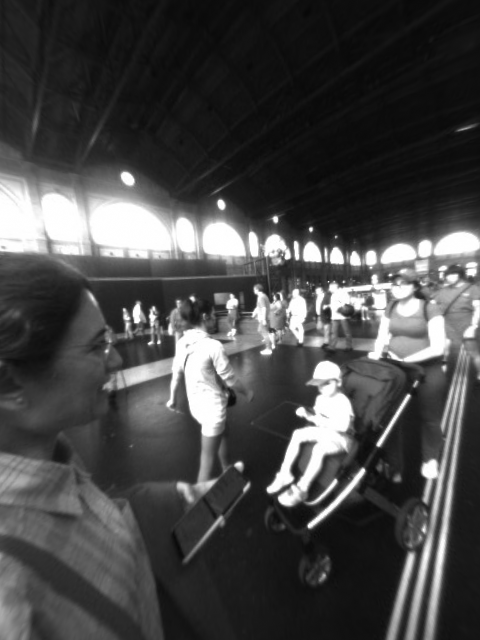}
        \end{minipage}\hfill
        \begin{minipage}[c]{0.33\textwidth}
            \centering
            \includegraphics[width=\textwidth, height=0.33\textheight,keepaspectratio]{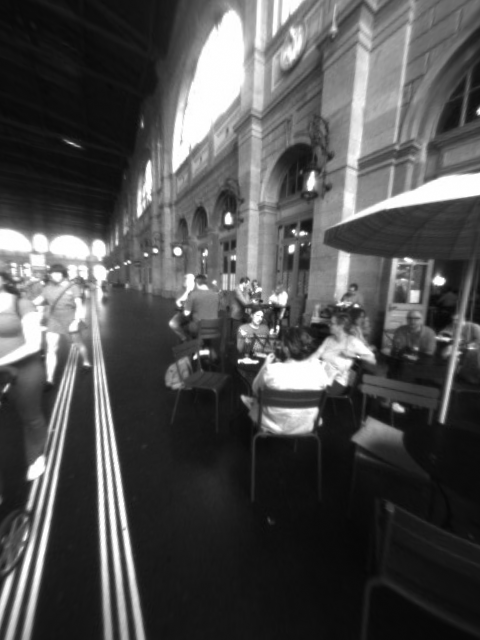}
        \end{minipage}
        
        \vspace{0.5em}

        \begin{minipage}[b]{\textwidth}
            \centering
            \includegraphics[width=0.96\textwidth]{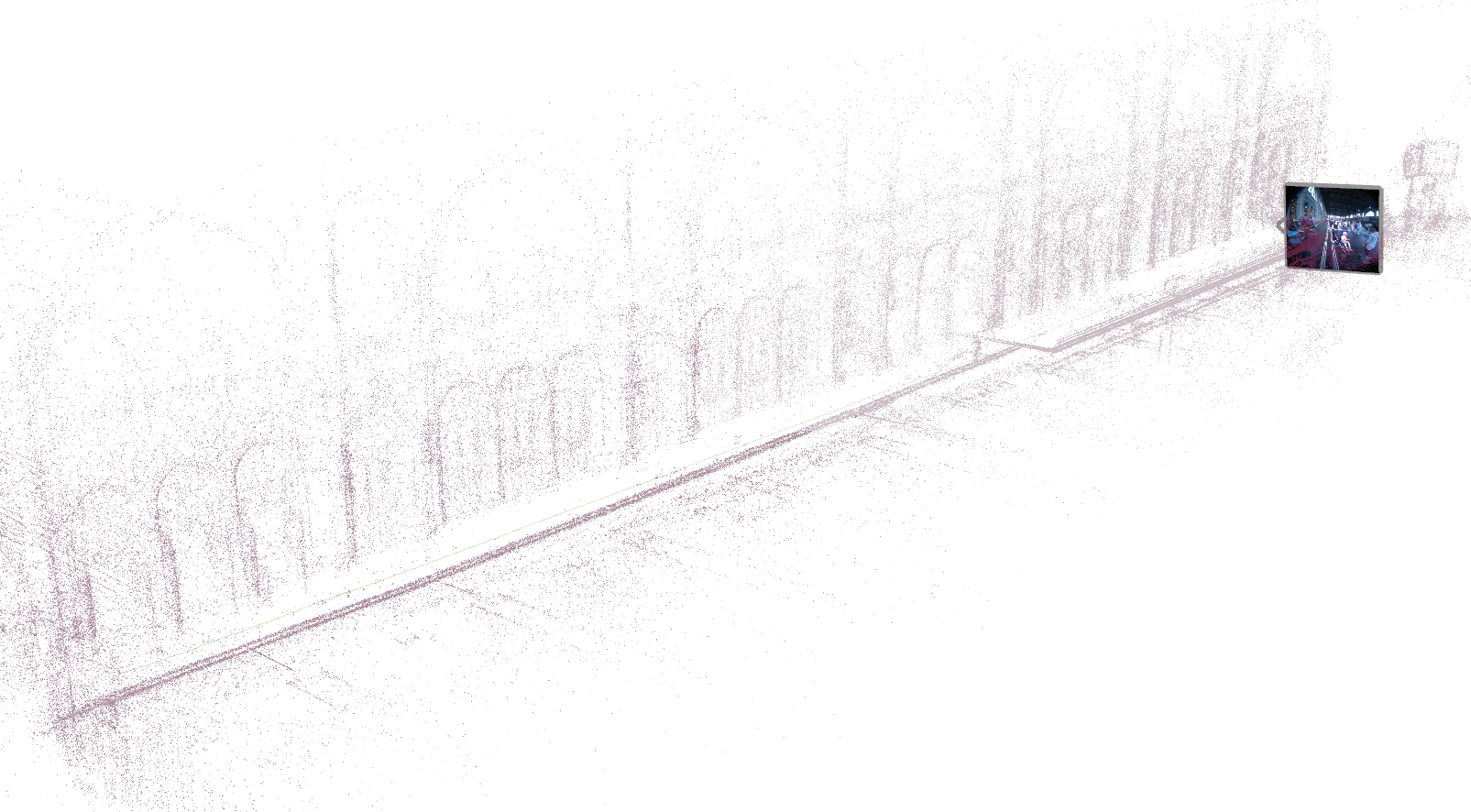}
        \end{minipage}
    \end{minipage}
}
\caption{Example of image data and corresponding 3D map from a dataset sequence: The top-left image shows the RGB frame, and the top-middle and top-right images represent the left and right images of a stereo pair. The bottom image shows the 3D map of the scene.\label{fig:example_camera_images_map}}
\end{figure}

Each dataset sequence in InCrowd-VI includes time\-stamped RGB images at resolutions of 1408 × 1408 and 30 FPS, stereo pair images at resolutions of 640 × 480 and 30 FPS, two streams of IMU data at data rates of 1000 and 800 Hz, semi-dense point clouds, and accurate trajectory ground truth. Although Meta-Ariana glasses provide data from multiple sensors, not all of them are included in the dataset, as they were not essential for the focus of this dataset. An example of the image data and their relative 3D map of the scene is shown in Figure~\ref{fig:example_camera_images_map}.
The dataset sequences are organized according to different levels of pedestrian density to facilitate a thorough evaluation of the SLAM algorithms. These levels are categorized as follows: High (more than 10 pedestrians per frame), Medium (4 to 10 pedestrians per frame), Low (1 to 3 pedestrians per frame), and None (no pedestrians present in the scene). In some sequences, severe occlusion caused by overcrowding is also present. Overcrowding occurs when pedestrians occupy the camera frame to the extent that a significant portion of the scene is obscured. This situation is characterized by substantial visual obstructions, where pedestrians temporarily block more than 50\% of the camera field of view. Furthermore, the sequences also consider the complexity of the environment.

\begin{figure}[H]
  \centering
   \includegraphics[width=0.8\linewidth]{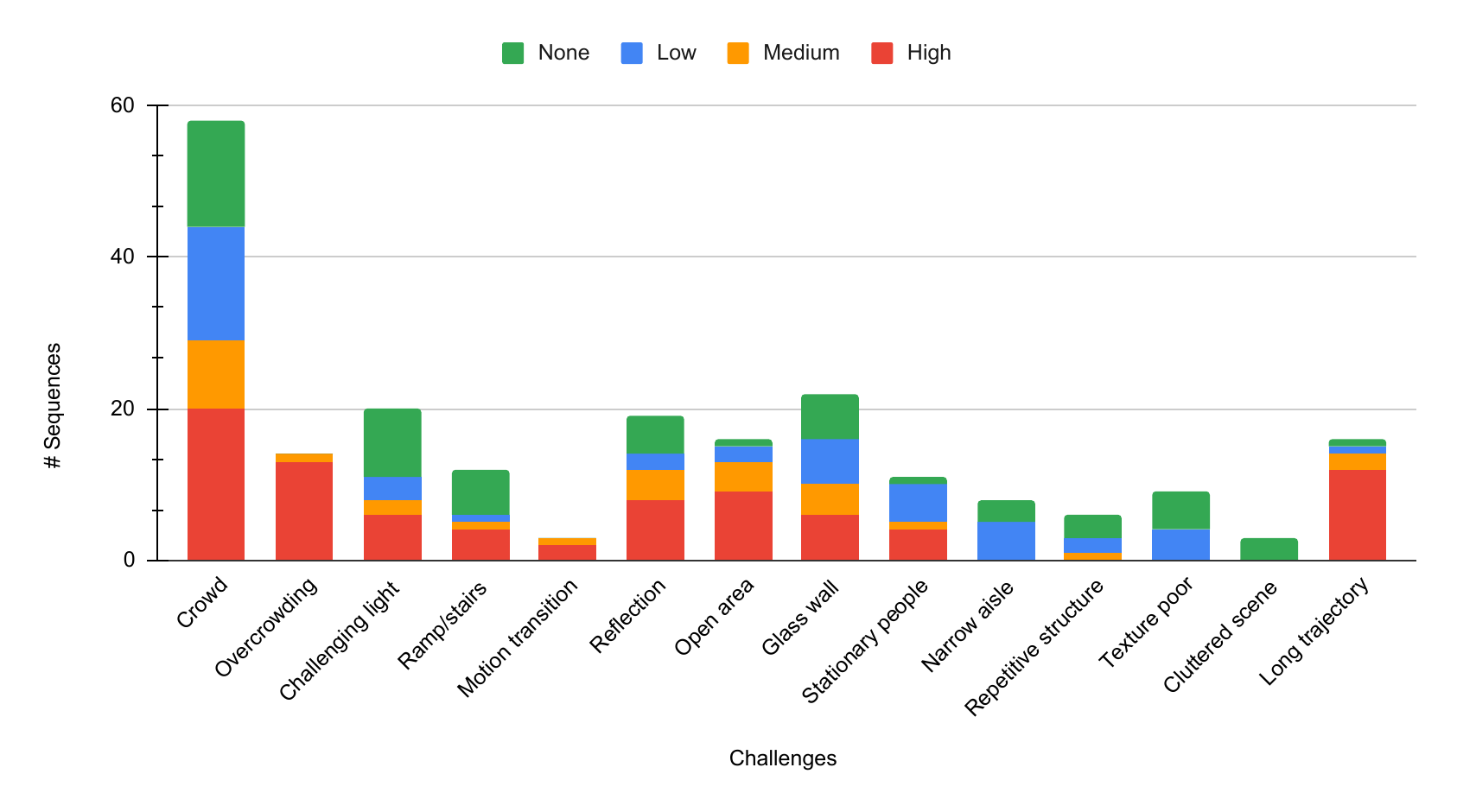}
   \caption{ Distribution of challenges across sequences in the InCrowd-VI dataset, categorized by crowd density levels (High: >10 pedestrians per frame, Medium: 4-10 pedestrians, Low: 1-3 pedestrians, None: no pedestrians). The x-axis represents the different types of challenges, and the y-axis indicates the total number of sequences. Note that the sequences may contain multiple challenges simultaneously.}
   \label{fig:challengs_statistics}
\end{figure}

\begin{figure}[H]
  \centering
   \includegraphics[width=0.8\linewidth]{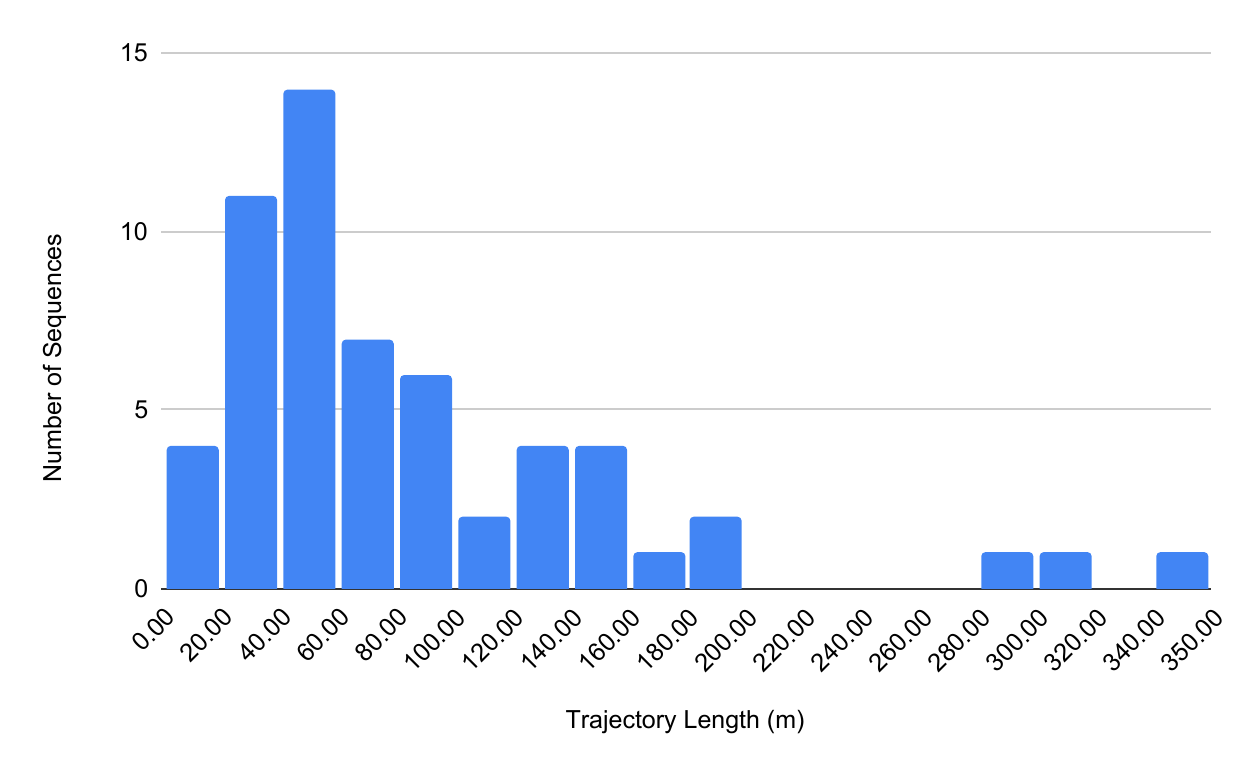}
   \caption{Histogram of trajectory length}
   \label{fig:trajectory_length}
\end{figure}

This categorization allows the evaluation of SLAM systems at varying levels of human density and environmental complexity. Table~\ref{tab:InCrowd-VI_sequences} provides an overview of the sequences, illustrating their categorization based on pedestrian density, diversity of venues, sequence lengths, duration, and the specific challenges encountered in each scenario. Table \ref{tab:InCrowd-VI_challenges_checkmarks} presents a detailed breakdown of the specific challenges in each sequence. The distribution of these challenges across different density levels enables comprehensive testing of SLAM systems under various combinations of human presence and environmental complexity.
Figure \ref{fig:challengs_statistics} shows the distribution of various challenges, broken down by crowd density levels. The visualization reveals that reflective surfaces and challenging lighting conditions are among the most common environmental challenges, reflecting typical indoor navigation scenarios.
In addition, the trajectory length plays a crucial role in assessing the performance and robustness of visual SLAM systems. For indoor environments for visually impaired navigation, we considered sequences with lengths shorter than 40 m as short trajectories, those ranging from 40 to 100 m as medium trajectories, and those of 100 m and beyond as long trajectories. Figure~\ref{fig:trajectory_length} shows a histogram of the trajectory lengths for the InCrowd-VI dataset.
\begin{table}[htbp!]
\caption{InCrowd-VI sequences. Density categories are defined as: High: >10 pedestrians per frame; Medium: 4-10 pedestrians per frame; Low: 1-3 pedestrians per frame; None: No pedestrians present in the scene. Sequences marked with an asterisk (*) in the "Sequence name" column include severe occlusion caused by overcrowding.}\label{tab:InCrowd-VI_sequences}
\begin{adjustwidth}{-\extralength}{0cm}
	\newcolumntype{L}{>{\arraybackslash}X}
    \begin{tabularx}{\fulllength}{llcccL}
    \toprule
     \textbf{Density} & \textbf{Sequence name} & \textbf{Venue} & \textbf{Length (m)} & \textbf{Span (mm:ss)} & \textbf{Main Challenges} \\ 
     \midrule
\multirow{20}{*}{\rotatebox{90}{\textbf{High}}} 
 & Oerlikon\_G7 & Train station & 133.36 & 02:04 & Challenging light, ramp \\
 & G6\_exit & Train station & 60.21 & 01:27 & Stairs, flickering light \\ 
 & G6\_loop * & Train station & 103.74 & 02:07 & Challenging light \\ 
 & Getoff\_pharmacy * & Train station & 159.50 & 03:06 & Motion transition \\ 
 & Service\_center & Airport & 131.89 & 02:18 & Reflection, open area \\ 
 & Ochsner\_sport * & Airport & 70.57 & 01:14 & Reflection \\ 
 & Toward\_gates & Airport & 165.46 & 02:57 & Reflection, open area \\ 
 & Arrival2 * & Airport & 74.96 & 01:23 & Crowd \\ 
 & Checkin2\_loop & Airport & 348.84 & 05:22 & Long trajectory \\ 
 & Shopping\_open * & Shopping mall & 182.51 & 03:10 & Challenging light, open area \\ 
 & Turn180 & Airport & 41.95 & 00:46 & Reflection \\ 
 & Cafe\_exit * & Main train station & 121.40 & 01:51 & Reflection, open area \\ 
 & G8\_cafe * & Main train station & 294.05 & 04:14 & Long trajectory \\ 
 & Ground\_53 * & Main train station & 32.40 & 01:01 & Escalator (motion transition) \\ 
 & Kiko\_loop * & Shopping mall & 314.68 & 05:50 & Long trajectory \\ 
 & Reservation\_office & Main train station & 95.15 & 01:28 & Reflection, open area \\ 
 & Orell\_fussli * & Shopping mall & 57.98 & 01:16 & Crowd \\ 
 & Reservation\_G17 * & Main train station & 97.30 & 01:50 & Challenging light, open area \\ 
 & Short\_Loop\_BH * & Main train station & 104.94 & 02:29 & Reflection, open area \\ 
 & Shopping\_loop * & Shopping mall & 151.91 & 02:36 & Challenging light \\ 
\midrule
\multirow{9}{*}{\rotatebox{90}{\textbf{Medium}}}
 & UZH\_stairs * & Museum & 13.16 & 00:41 & Stairs \\ 
 & Migros & Shopping mall & 84.18 & 01:40 & Reflection, open area \\
 & TS\_exam\_loop & Exam location & 28.81 & 01:34 & Glass wall, stationary people \\ 
 & Museum\_loop & Museum & 66.99 & 02:02 & Challenging light \\ 
 & Ramp\_checkin2 & Airport & 191.77 & 03:40 & Moving ramp, open area \\ 
 & Airport\_shop & Shopping mall & 99.49 & 01:46 & Reflection, open area \\ 
 & Towards\_checkin1 & Airport & 55.89 & 01:02 & Reflection, open area, glass wall \\ 
 & Entrance\_checkin1 & Airport & 35.70 & 00:39 & Reflection \\ 
 & Towards\_circle & Airport & 127.46 & 02:06 & Challenging light, repetitive structure \\ 
\midrule
\multirow{15}{*}{\rotatebox{90}{\textbf{Low}}} 
 & AND\_floor51 & Library & 40.23 & 00:53 & Narrow aisle, stationary people \\ 
 & AND\_floor52 & Library & 39.13 & 01:10 & Narrow aisle, stationary people \\ 
 & AND\_liftAC & University building & 71.15 & 01:33 & Open area, glass wall \\ 
 & ETH\_HG & University building & 99.00 & 01:56 & Repetitive structure \\ 
 & ETH\_lab & Laboratory & 56.70 & 01:24 & Reflection \\ 
 & Kriegsstr\_pedestrian & Public building & 31.97 & 00:59 & Texture-poor \\ 
 & Kriegsstr\_same\_dir & Public building & 31.17 & 00:56 & Texture-poor \\
 & TH\_entrance & University building & 41.85 & 00:53 & Challenging light, stationary people \\ 
 & TS\_entrance & University building & 32.06 & 00:42 & Reflection \\ 
 & TS\_exam & University building & 44.93 & 01:01 & Narrow corridor, texture-poor \\ 
 & UZH\_HG & University building & 142.12 & 03:09 & Long trajectory, repetitive structure \\ 
 & Museum\_1 & Museum & 69.30 & 01:51 & Challenging light \\ 
 & Museum\_dinosaur & Museum & 44.62 & 01:16 & Challenging light \\ 
 & Museum\_up & Museum & 12.20 & 00:34 & Stairs \\ 
 & Short\_loop & Airport & 36.92 & 00:46 & Open area \\ 
\midrule
\multirow{14}{*}{\rotatebox{90}{\textbf{None}}}
 & AND\_Lib & Office building & 52.19 & 01:18 & Reflection, narrow corridor \\ 
 & Hrsaal1B01 & Academic building & 74.55 & 01:50 & Challenging light, narrow corridor \\ 
 & ETH\_FT2 & Museum & 40.97 & 00:56 & Challenging light, reflection \\ 
 & ETH\_FTE & Museum & 59.00 & 01:35 & Challenging light, open area \\ 
 & ETH\_lab2 & Laboratory & 58.08 & 01:26 & Texture-poor, reflection \\ 
 & Habsburgstr\_dark & Public building & 36.08 & 01:05 & Stairs, dimly lit \\ 
 & Habsburgstr\_light & Public building & 87.99 & 02:46 & Stairs \\ 
 & IMS\_lab & Laboratory & 15.23 & 00:43 & Cluttered scene \\ 
 & IMS\_TE21 & Laboratory & 42.43 & 01:23 & Cluttered scene, challenging light \\ 
 & IMS\_LEA & Laboratory & 19.35 & 00:43 & Cluttered scene \\ 
 & Kriegsstr & Public building & 31.18 & 00:54 & Texture-poor \\ 
 & TH\_loop & Office building & 154.74 & 03:55 & Reflection \\
 & TS116 & University building & 59.11 & 01:24 & Challenging light, reflection, glass wall \\ 
 & TS\_stairs & University building & 27.67 & 00:52 & Stairs, challenging light, glass wall \\ 
			\bottomrule
		\end{tabularx}
	\end{adjustwidth}
\end{table}

\begin{table}[htbp!]
\centering
\scriptsize
\caption{Challenges present in each sequence in InCrowd-VI. A checkmark (\checkmark) indicates the presence of a specific challenge in the sequence.}
\label{tab:InCrowd-VI_challenges_checkmarks}
\begin{tabular}{llccccccccccccc}
\toprule
\textbf{Density} & \textbf{Sequence} & \rotatebox{90}{\textbf{Overcrowding}} & \rotatebox{90}{\textbf{Challenging light}} & \rotatebox{90}{\textbf{Ramp/Stairs}} & \rotatebox{90}{\textbf{Motion transition}} & \rotatebox{90}{\textbf{Reflection}} & \rotatebox{90}{\textbf{Open area}} & \rotatebox{90}{\textbf{Glass wall}} & \rotatebox{90}{\textbf{Stationary people}} & \rotatebox{90}{\textbf{Narrow aisle/corridor}} & \rotatebox{90}{\textbf{Repetitive structure}} & \rotatebox{90}{\textbf{Texture poor}} & \rotatebox{90}{\textbf{Cluttered scene}} & \rotatebox{90}{\textbf{Long trajectory}} \\ 
\midrule
\multirow{20}{*}{\rotatebox{90}{\textbf{High}}} 
 & Oerlikon\_G7 & & \checkmark & \checkmark & & & & & & & & & \checkmark \\ 
 & G6\_exit & & \checkmark & \checkmark & & & & & & & & & & \\ 
 & G6\_loop & \checkmark & \checkmark & & & & & \checkmark & & & & & & \checkmark \\ 
 & Getoff\_pharmacy & \checkmark & & \checkmark & \checkmark & & & \checkmark & & & & & & \checkmark \\ 
 & Service\_center & & & & & \checkmark & \checkmark & \checkmark & & & & & & \checkmark \\ 
 & Ochsner\_sport & \checkmark & & & & \checkmark & & \checkmark & & & & & & \\ 
 & Toward\_gates & & & \checkmark & & \checkmark & \checkmark & & & & & & & \checkmark \\ 
 & Arrival2 & \checkmark & & & & & & \checkmark & \checkmark & & & & & \\ 
 & Checkin2\_loop & & & & & \checkmark & \checkmark & & \checkmark & & & & & \checkmark \\ 
 & Shopping\_open & \checkmark & \checkmark & & & & \checkmark & & & & & & & \checkmark \\ 
 & Turn180 & & & & & \checkmark & & & & & & & & \\ 
 & Cafe\_exit & \checkmark & & & & \checkmark & \checkmark & & \checkmark & & & & & \checkmark \\ 
 & G8\_cafe & \checkmark & & & & & & & \checkmark & & & & & \checkmark \\ 
 & Ground\_53 & \checkmark & & & \checkmark & & & & & & & & & \\ 
 & Kiko\_loop & \checkmark & & & & & & & & & & & & \checkmark \\ 
 & Reservation\_office & & & & & \checkmark & \checkmark & & & & & & & \\ 
 & Orell\_fussli & \checkmark & & & & & & & & & & & & \\ 
 & Reservation\_G17 & \checkmark & \checkmark & & & & \checkmark & & & & & & & \\ 
 & Short\_Loop\_BH & \checkmark & & & & \checkmark & \checkmark & & \checkmark & & & & & \checkmark \\ 
 & Shopping\_loop & \checkmark & \checkmark & & & & \checkmark & \checkmark & & & & & & \checkmark \\ 
\midrule
\multirow{9}{*}{\rotatebox{90}{\textbf{Medium}}}
 & UZH\_stairs & \checkmark & & \checkmark & & & & & & & & & & \\ 
 & Migros & & & & & \checkmark & \checkmark & \checkmark & & & & & & \\ 
 & TS\_exam\_loop & & & & & & & \checkmark & \checkmark & & & & & \\ 
 & Museum\_loop & & \checkmark & & & & & \checkmark & & & & & & \\ 
 & Ramp\_checkin2 & & & \checkmark & \checkmark & & \checkmark & & & & & & & \checkmark \\ 
 & Airport\_shop & & & & & \checkmark & \checkmark & & & & & & & \\ 
 & Towards\_checkin1 & & & & & \checkmark & \checkmark & \checkmark & & & & & & \\ 
 & Entrance\_checkin1 & & & & & \checkmark & & & & & & & & \\ 
 & Towards\_circle & & \checkmark & & & & & & & & \checkmark & & & \checkmark \\ 
\midrule
\multirow{15}{*}{\rotatebox{90}{\textbf{Low}}} 
 & AND\_floor51 & & & & & & & & \checkmark & \checkmark & & & & \\ 
 & AND\_floor52 & & & & & & & & \checkmark & \checkmark & & & & \\ 
 & AND\_liftAC & & & & & & \checkmark & \checkmark & & & & & & \\ 
 & ETH\_HG & & & & & & & & & & \checkmark & & & \\ 
& ETH\_lab & & & & & \checkmark & & \checkmark & & & & & & \\ 
 & Kriegsstr\_pedestrian & & & & & & & & & \checkmark & & \checkmark & & \\ 
 & Kriegsstr\_same\_dir & & & & & & & & & \checkmark & & \checkmark & & \\ 
 & TH\_entrance & & \checkmark & & & & & \checkmark & \checkmark & & & & & \\ 
 & TS\_entrance & & & & & \checkmark & & \checkmark & & & & & & \\ 
 & TS\_exam & & & & & & & & \checkmark & \checkmark & & \checkmark & & \\ 
 & UZH\_HG & & & & & & & & \checkmark & & \checkmark & \checkmark & & \checkmark \\ 
 & Museum\_1 & & \checkmark & & & & & \checkmark & & & & & & \\ 
 & Museum\_dinosaur & & \checkmark & & & & & \checkmark & & & & & & \\ 
 & Museum\_up & & & \checkmark & & & & & & & & & & \\ 
 & Short\_loop & & & & & & \checkmark & & & & & & & \\ 
\midrule
\multirow{14}{*}{\rotatebox{90}{\textbf{None}}}
 & AND\_Lib & & & & & \checkmark & & & & \checkmark & & & & \\ 
 & Hrsaal1B01 & & \checkmark & & & & & \checkmark & & \checkmark & \checkmark & & & \\ 
 & ETH\_FT2 & & \checkmark & & & \checkmark & & & & & \checkmark & & & \\ 
 & ETH\_FTE & & \checkmark & & & & \checkmark & & & & & & & \\ 
 & ETH\_lab2 & & & & & \checkmark & & \checkmark & & & & \checkmark & & \\ 
 & Habsburgstr\_dark & & \checkmark & \checkmark & & & & & & & & \checkmark & & \\ 
 & Habsburgstr\_light & & \checkmark & \checkmark & & & & & & & & \checkmark & & \\ 
 & IMS\_lab & & & & & & & & & & & & \checkmark & \\ 
 & IMS\_TE21 & & \checkmark & \checkmark & & & & & \checkmark & \checkmark & & & \checkmark & \\ 
 & IMS\_LEA & & & & & & & & & & & & \checkmark & \\ 
 & Kriegsstr & & & & & & & & & \checkmark & & \checkmark & & \\ 
 & TH\_loop & & & & & \checkmark & & \checkmark & & & & & & \checkmark \\ 
 & TS116 & & \checkmark & & & \checkmark & & \checkmark & & & & \checkmark & & \\ 
 & TS\_stairs & & \checkmark & \checkmark & & & & \checkmark & & & \checkmark & & \\ 
\bottomrule
\end{tabular}
\end{table}
It is important to note that the walking speeds in our dataset reflect the typical navigation patterns of visually impaired individuals. Studies have shown that people with visual impairment tend to walk slower and have reduced stride lengths during independent and guided walking compared to sighted people \cite{bennett2019walking}. While the typical walking speed for sighted populations is generally between 1.11 and 1.4 m/s \cite{bennett2019walking}, the average walking speed in the InCrowd-VI dataset is 0.75 m/s. This lower average speed aligns with the expectations of visually impaired navigation.

\begin{figure}[H]
  \centering
    \begin{minipage}{\textwidth}
      \centering
      \begin{subfigure}{0.32\linewidth}
        \includegraphics[width=\textwidth]{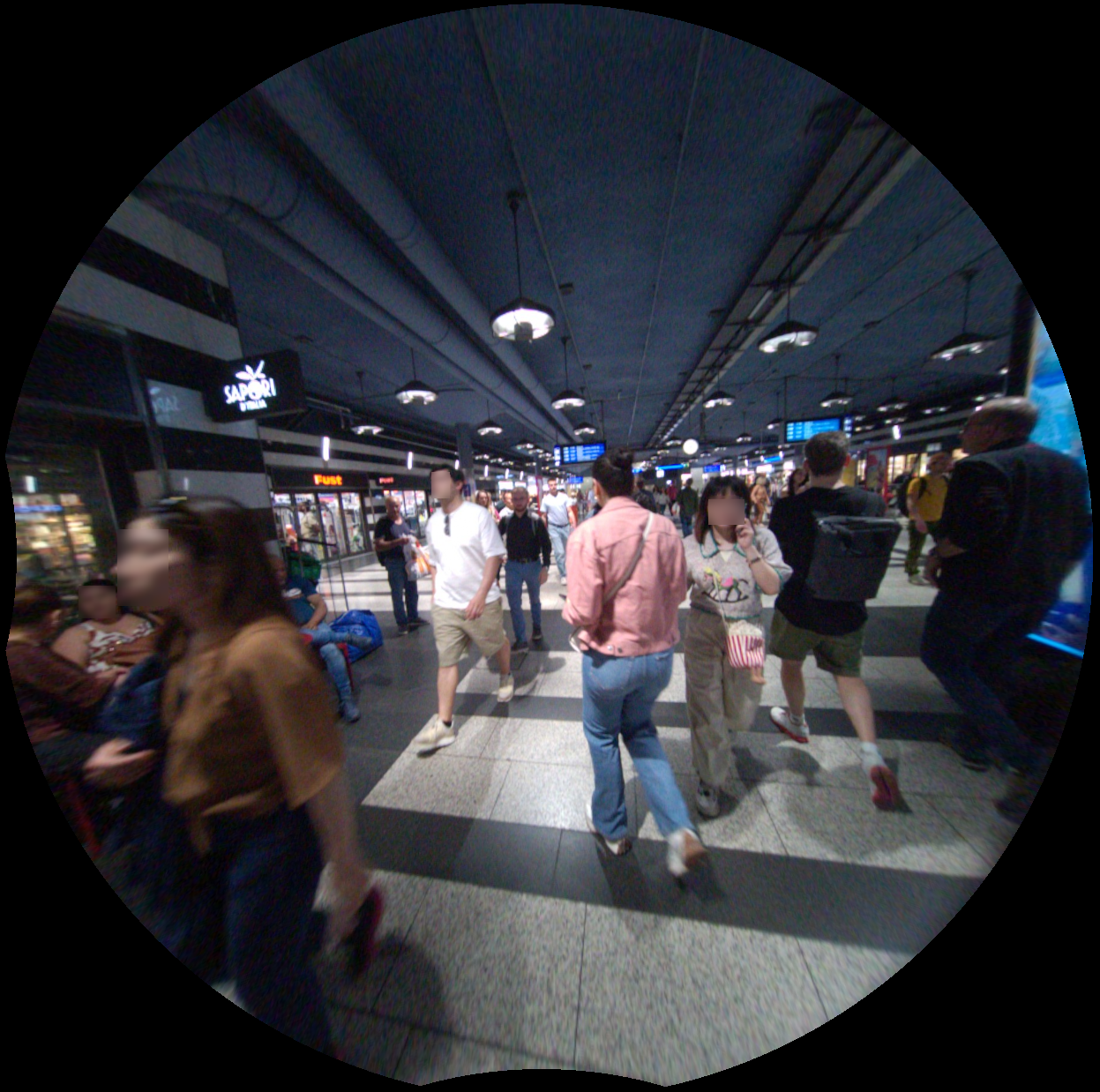}
        \caption{}
        \label{fig:subfig1}
      \end{subfigure}
      \hfill
      \begin{subfigure}{0.32\textwidth}
        \includegraphics[width=\textwidth]{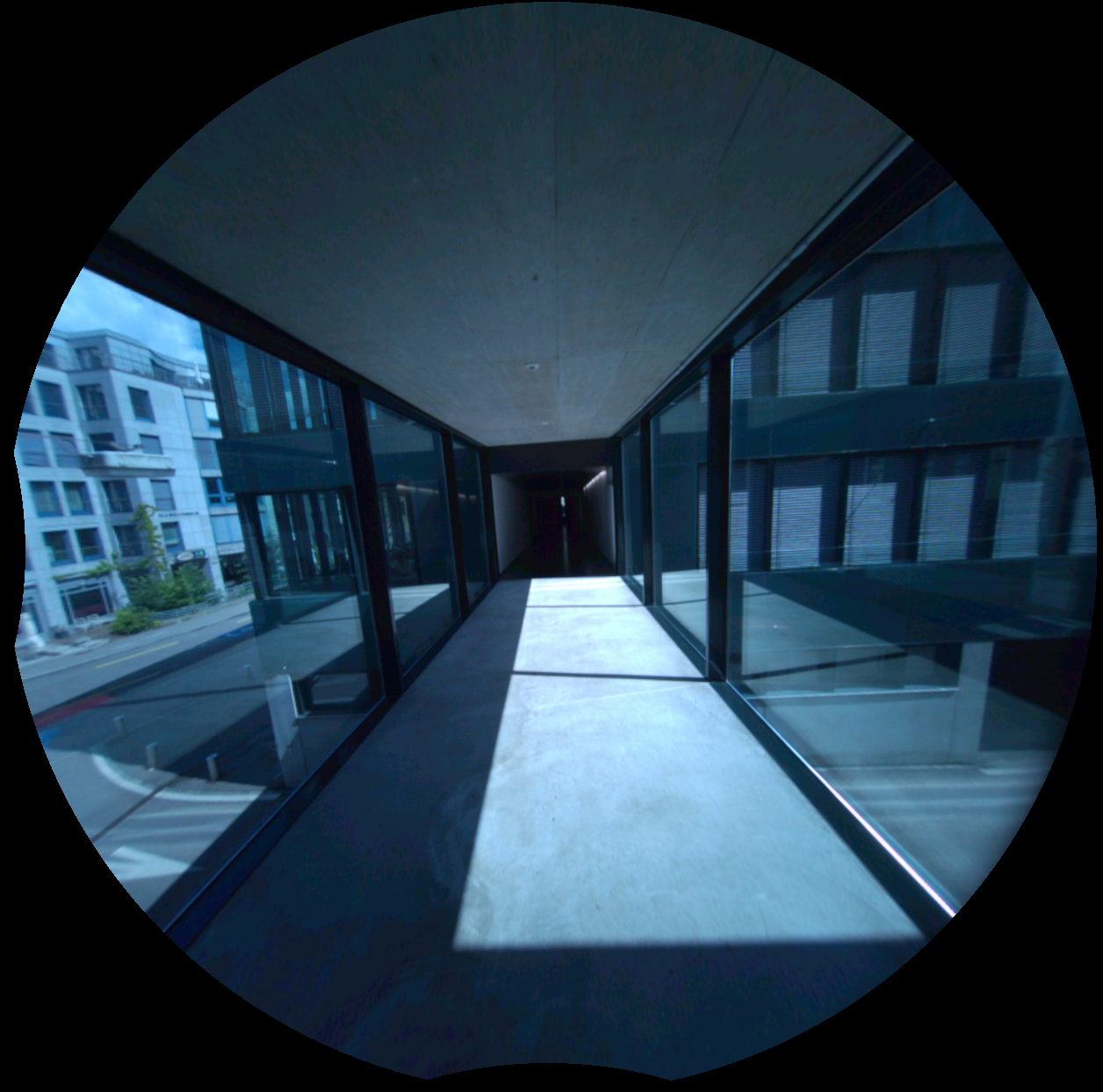}
        \caption{}
        \label{fig:subfig2}
      \end{subfigure}
      \hfill
      \begin{subfigure}{0.32\textwidth}
        \includegraphics[width=\textwidth]{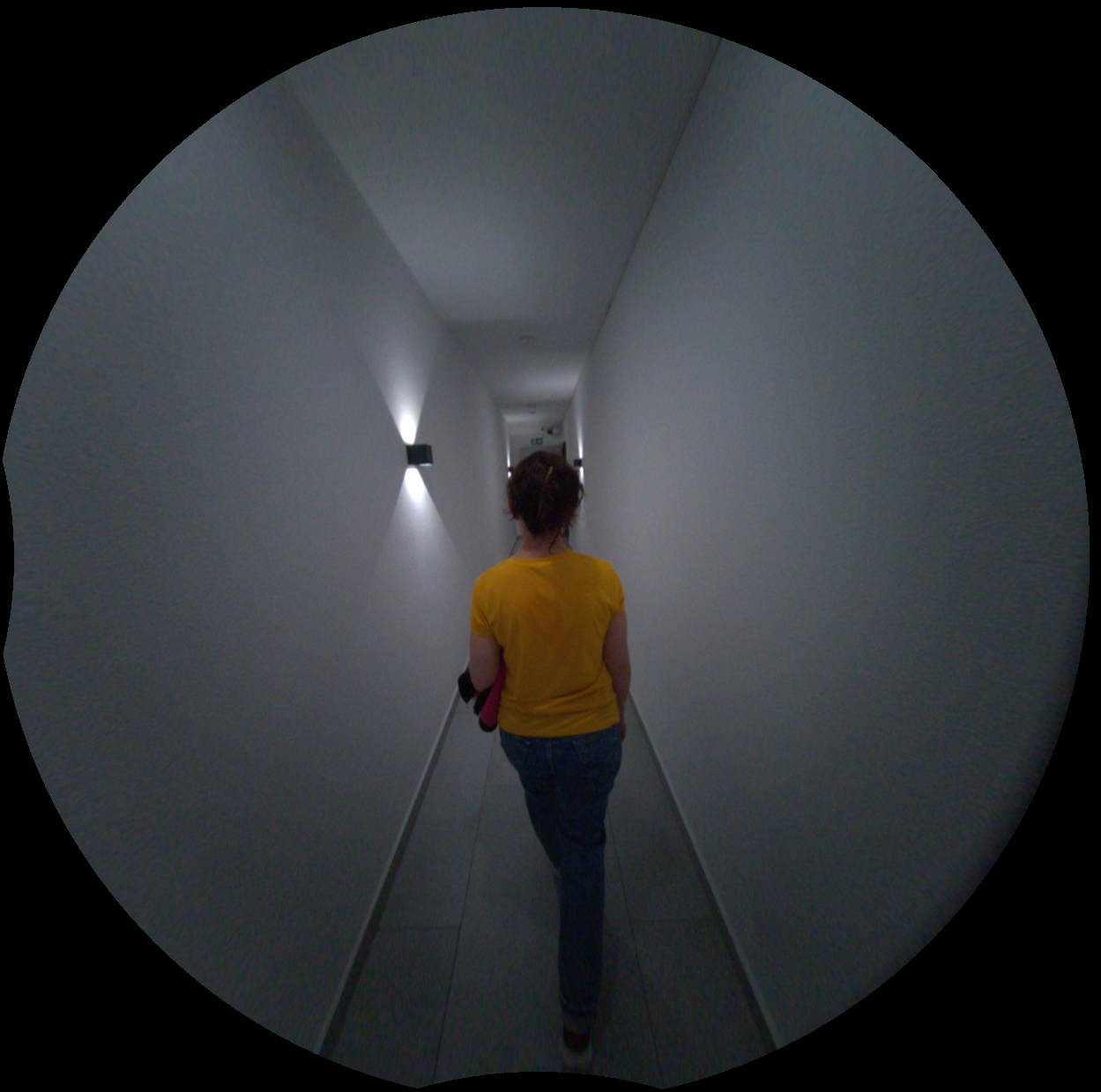}
        \caption{}
        \label{fig:subfig3}
      \end{subfigure}

      \vspace{0.3cm} 
      
      \begin{subfigure}{0.32\textwidth}
        \includegraphics[width=\textwidth]{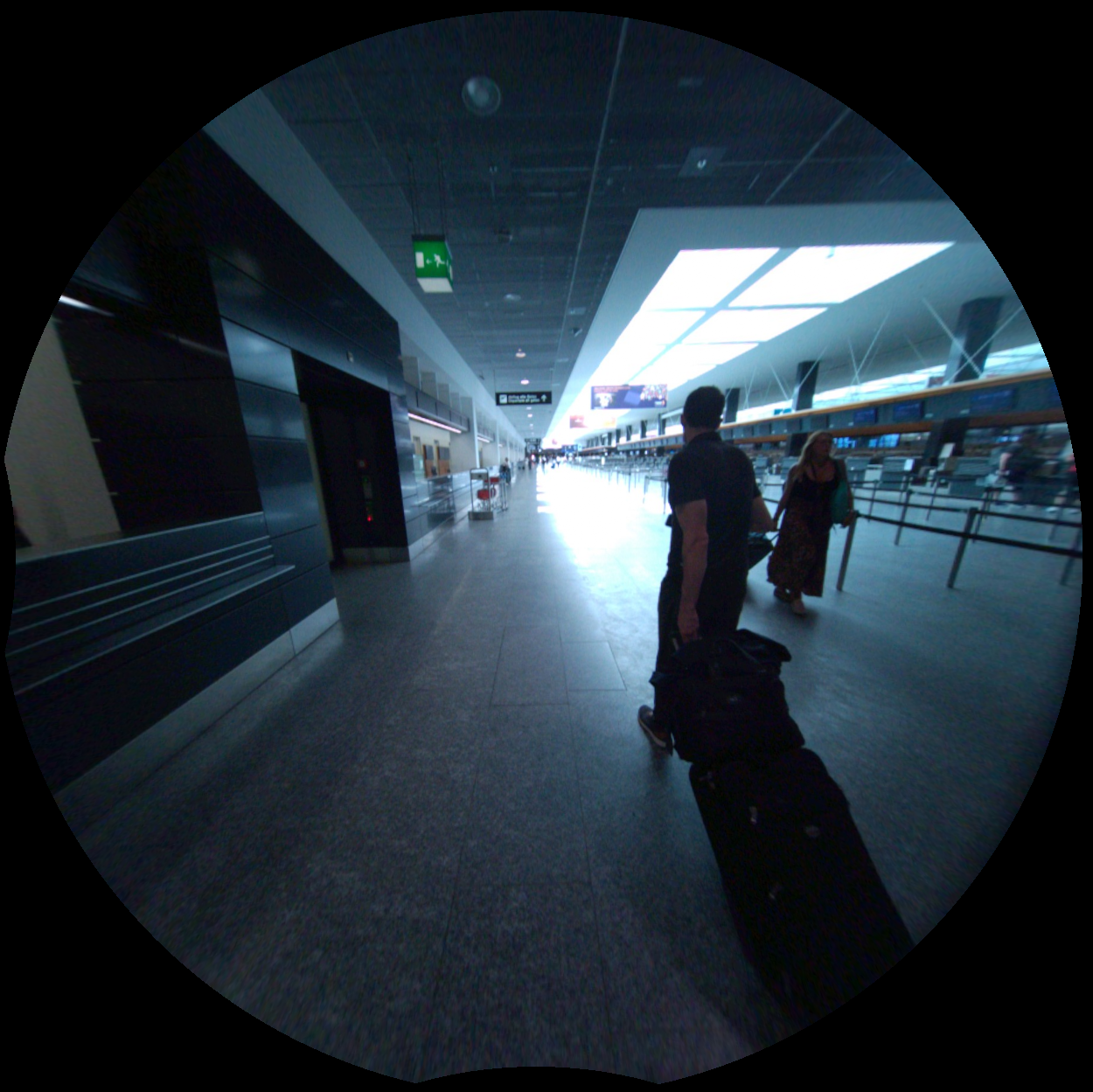}
        \caption{}
        \label{fig:subfig4}
      \end{subfigure}
      \hfill
      \begin{subfigure}[b]{0.32\textwidth}
        \includegraphics[width=\textwidth]{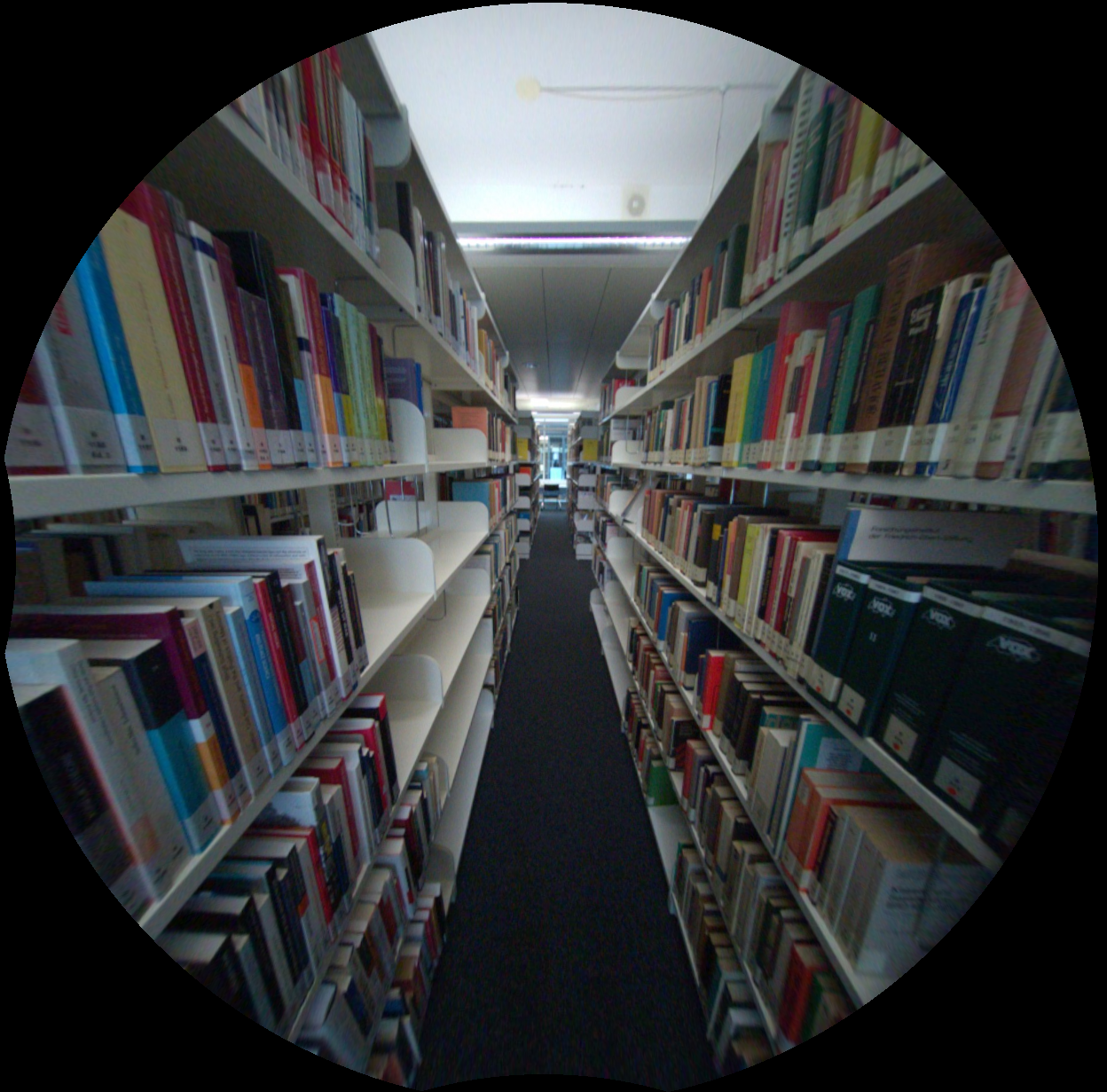}
        \caption{}
        \label{fig:subfig5}
      \end{subfigure}
      \hfill
      \begin{subfigure}[b]{0.32\textwidth}
        \includegraphics[width=\textwidth]{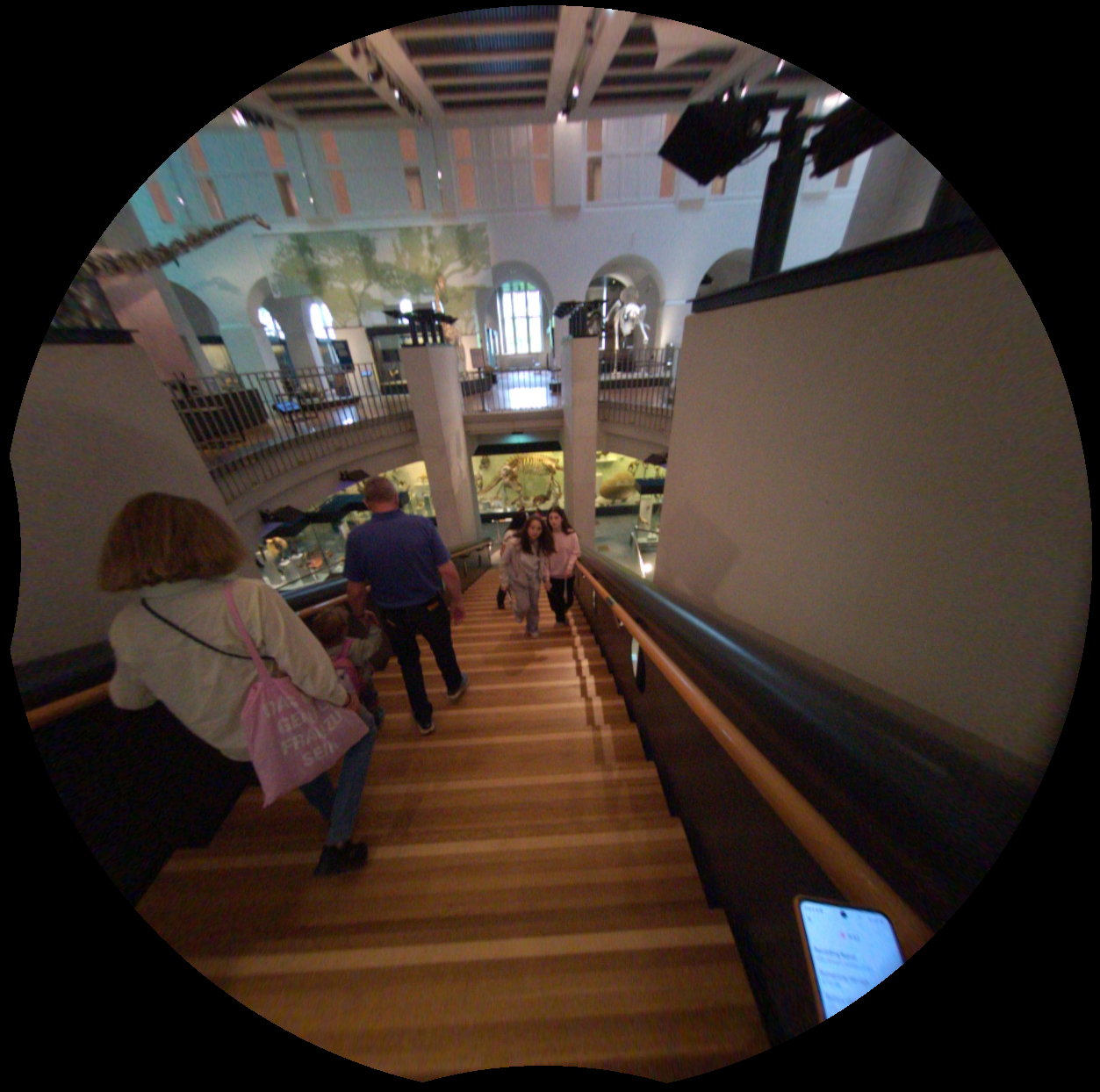}
        \caption{}
        \label{fig:subfig6}
      \end{subfigure}
    \end{minipage}
  \caption{Example scenes from the InCrowd-VI dataset demonstrating various challenges: (a) high pedestrian density, (b) varying lighting conditions, (c) texture-poor surfaces, (d) reflective surfaces, (e) narrow aisles, (f) stairs.\label{fig:dataset_challenges}}
\end{figure}

The dataset captures a wide range of challenges inherent in real-world indoor navigation scenarios, including the following.
\begin{itemize}
    \item Dynamic obstacles: InCrowd-VI features sequences with moving pedestrians, capturing scenarios of crossing paths with groups and maneuvering around individuals moving in different directions. These sequences test the ability of the SLAM systems to handle unpredictable dynamic elements in real-world environments.
    \item Crowd density variation: Sequences capture a range of crowd densities, from static to densely populated areas, testing the adaptability of SLAM systems to different levels of human activity.
    \item Frequent occlusions: The dataset includes sequences with frequent occlusions caused by moving pedestrians, luggage, and infrastructure, thereby creating significant challenges for maintaining accurate mapping and tracking.
    \item Reflective and transparent surfaces: The dataset includes scenes with glass and other reflective surfaces that can distort sensor readings and complicate the visual SLAM algorithms.
    \item Texture-poor areas: Scenes with minimal visual features, such as plain walls, challenge feature-based SLAM systems.
    \item Large-scale and complex environments: The dataset covers diverse architectural layouts, including open spaces, corridors, ramps, staircases, and escalators, to test the adaptability of SLAM to various spatial configurations.
    \item Lighting variations:  Sequences incorporate sequences with varying lighting conditions, from well-lit atriums to dimly lit corridors or areas with flickering lights, to test the SLAM robustness under varying illumination conditions.
    \item Sudden viewpoint changes: Sequences capture user perspective shifts during corner turns and level transitions, thereby challenging SLAM tracking consistency.
    \item Motion transitions: Sequences include transitions between moving environments (escalators, moving ramps, and trains) and stationary areas, to test SLAM's ability to distinguish ego-motion from environmental motion.
\end{itemize}
These challenges collectively contribute to the realism and complexity of the InCrowd-VI dataset, making it a valuable resource for evaluating and advancing SLAM systems designed for visually impaired navigation in real-world indoor environments. Figure~\ref{fig:dataset_challenges} shows the selection of images from the InCrowd-VI dataset.
\section{Experimental Evaluation}
\label{sec:experimental_evaluation}
To assess the accuracy and robustness of current state-of-the-art visual SLAM systems on the InCrowd-VI dataset, four representative algorithms were selected: two classical approaches and two deep learning-based methods. These systems were chosen because of their prominence in the field and diverse approaches to visual SLAM. As shown in Table~\ref{tab:slam_comparison}, DROID-SLAM \cite{teed2021droid} and DPV-SLAM \cite{lipson2024deep} use dense optical flow-based and sparse patch-based matching approaches, respectively, both of which leverage end-to-end learning. Conversely, ORB-SLAM3 \cite{campos2021orb} and SVO \cite{forster2016svo} employ the classical feature matching and semi-direct matching approaches, respectively. This evaluation serves to demonstrate that InCrowd-VI effectively captures challenging real-world scenarios that current state-of-the-art systems struggle to handle and underscores the dataset's value as a benchmark for advancing SLAM research in visually impaired navigation.  
\begin{table}
  \centering
  \footnotesize
  \begin{tabular}{@{}lll@{}}
    \toprule
    SLAM system & Matching approach & Approach \\
    \midrule
    DROID-SLAM & Dense optical flow-based & End-to-end \\
    DPV-SLAM & Sparse patch-based & End-to-end \\
    ORB-SLAM3 & Feature-based & Classical SLAM \\
    SVO & Semi-direct & Classical VO \\
    \bottomrule
  \end{tabular}
  \caption{Characteristics of selected SLAM systems}
  \label{tab:slam_comparison}
\end{table}

A selection of sequences from InCrowd-VI representing a range of difficulty levels from easy to hard was used for evaluation. These sequences were selected to represent the diverse challenges present in the dataset, including changing elevations, long trajectories, motion transitions, challenging lighting conditions, and crowds with different densities. This approach allows for a comprehensive assessment of the accuracy and robustness of SLAM systems across various scenarios, while maintaining a manageable evaluation process. ORB-SLAM3 was evaluated using the left camera images and left IMU measurements. The DROID-SLAM, DPV-SLAM, and SVO were evaluated using only the left camera images.
Additionally, the intrinsic camera parameters were extracted, and the left IMU-to-left camera transformation was calculated using the calibration data provided by the Meta Aria project. For camera calibration, Project Aria used a sophisticated camera model (FisheyeRadTanThinPrism) \footnote{https://facebookresearch.github.io/projectaria\_tools/docs/tech\_insights/camera\_intrinsic\_models}, which includes six radial, two tangential, and four thin-prism distortion parameters. A custom tool was developed to convert the data from the VRS files recorded by the Meta Aria glasses to the required format for each VO and SLAM system, and system configuration parameter files were provided for each system to ensure compatibility with the dataset.
The experimental setup ran Ubuntu 20.04 and 22.04 on a Lenovo laptop equipped with a 12\textsuperscript{th} Gen Intel\textsuperscript{\textregistered} Core\texttrademark\ i9-12950HX vPro\textregistered\ Processor, 64 GB DDR5 RAM, and an NVIDIA RTX\texttrademark\ A2000 8 GB GDDR6 graphics card.
ORB-SLAM3, SVO, and DPV-SLAM incorporate random processes into their operations, such as RANSAC for outlier rejection and random feature selection. DPV-SLAM, which is based on DPVO \cite{teed2024deep}, adds further variability through random keypoint selection and its probabilistic depth estimation approach. By contrast, DROID-SLAM uses a deterministic model that produces identical results across multiple runs, as observed in our experiments. To account for these algorithmic differences and ensure fair evaluation, we executed ORB-SLAM3, SVO, and DPV-SLAM five times each, reporting mean values, whereas DROID-SLAM required only a single execution per sequence. The results are summarized in Table~\ref{tab:evaluation_results}.
\startlandscape
\begin{table}[H]
\caption{Quantitative results of evaluated SLAM systems across various challenging scenarios, including different crowd densities, lighting conditions, long trajectories, elevation changes, and motion transitions. The results include the absolute trajectory error (ATE) in meters, drift percentage (DP) showing system drift as a percentage of total trajectory length, pose estimation coverage (PEC) as a percentage, system frames per second (FPS) indicating the number of frames processed by the system per second, and real-time factor (RTF). RTF provides a ratio that indicates how much faster the system processes data compared to the user's walking speed. The table includes the trajectory length (Trj. Len.) in meters and the average walking speed (AWS) in meters per second for each sequence. Sequences with high crowd density are marked in red, medium in orange, low in blue, and none in black. The failed sequences are denoted as ×.\label{tab:evaluation_results}}

\begin{tabularx}{\textwidth}{l|c|c|ccccc|ccccc|ccccc|ccccc}
\toprule
    \multirow{3}{*}{Sequence} & \multirow{3}{*}{\makecell{Trj\\Len\\(m)}} & \multirow{3}{*}{\makecell{AWS\\(m/s)}} & \multicolumn{10}{c|}{Classical Systems} & \multicolumn{10}{c}{Deep Learning-Based Systems} \\
    \cline{4-23}
    & & & \multicolumn{5}{c|}{ORB-SLAM3} & \multicolumn{5}{c|}{SVO} & \multicolumn{5}{c|}{DROID-SLAM} & \multicolumn{5}{c}{DPV-SLAM} \\
    \cline{4-23}
    & & & \makecell{ATE\\(m)} & \makecell{DP\\(\%)} & \makecell{PEC\\(\%)} & FPS & RTF & \makecell{ATE\\(m)} & \makecell{DP\\(\%)} & \makecell{PEC\\(\%)} & FPS & RTF & \makecell{ATE\\(m)} & \makecell{DP\\(\%)} & \makecell{PEC\\(\%)} & FPS & RTF & \makecell{ATE\\(m)} & \makecell{DP\\(\%)} & \makecell{PEC\\(\%)} & FPS & RTF \\
    \midrule
    \multicolumn{23}{c}{\textbf{Crowd Density}} \\
    \midrule
    \textcolor{red}{Orell\_fussli} & 57.98 & 0.76 
        & 2.95 & 5.08 & 90 & 23 & 0.80 
        & x & x & x & 174 & 5.82 
        & 1.07 & 1.84 & 100 & 16 & 0.53 
        & 0.21 & 0.36 & 98 & 10 & 0.36 \\ 
    \textcolor{orange}{Entrance\_checkin1} & 35.70 & 0.92 
        & 0.77 & 2.15 & 92 & 25 & 0.82 
        & 6.78 & 18.99 & 80 & 128 & 4.27 
        & 0.08 & 0.22 & 100 & 16 & 0.53
        & 0.08 & 0.22 & 97 & 11 & 0.39 \\ 
    \textcolor{blue}{Short\_loop} & 36.92 & 0.80 
        & 0.22 & 0.59 & 96 & 24 & 0.83 
        & 3.71 & 10.04 & 70 & 134 & 4.51 
        & 0.04 & 0.10 & 100 & 15 & 0.51 
        & 0.32 & 0.86 & 97 & 11 & 0.38 \\ 
    IMS\_lab & 15.23 & 0.35 
        & 0.06 & 0.39 & 94 & 24 & 0.82 
        & 2.49 & 16.34 & 71 & 149 & 5.05 
        & 0.02 & 0.13 & 100 & 16 & 0.54 
        & 0.04 & 0.26 & 97 & 14 & 0.48 \\ 
    \hline
    \multicolumn{23}{c}{\textbf{Lighting Variations}} \\
    \hline
    \textcolor{red}{Reservation\_G17} & 97.30 & 0.88 
        & 7.45 & 7.65 & 95 & 26 & 0.87 
        & 15.23 & 15.65 & 74 & 156 & 5.25 
        & 1.14 & 1.17 & 100 & 19 & 0.65 
        & 1.95 & 2.00 & 99 & 12 & 0.40 \\ 
    \textcolor{orange}{Toward\_circle} & 127.46 & 1.01 
        & x & x & 97 & 25 & 0.86 
        & x & x & x & 175 & 5.86 
        & 0.88 & 0.69 & 100 & 15 & 0.50 
        & 0.59 & 0.46 & 99 & 10 & 0.29 \\ 
    \textcolor{blue}{Museum\_1} & 69.30 & 0.62 
        & 0.34 & 0.49 & 98 & 25 & 0.87 
        & 3.82 & 5.51 & 36 & 171 & 5.75 
        & 6.21 & 8.96 & 100 & 18 & 0.61 
        & 8.06 & 11.63 & 99 & 10 & 0.34 \\ 
    TS116 & 59.11 & 0.70 
        & x & x & 57 & 21 & 0.71 
        & 3.99 & 6.75 & 31 & 159 & 5.32 
        & 3.77 & 6.37 & 100 & 15 & 0.52 
        & 1.08 & 1.82 & 98 & 10 & 0.34 \\ 
    \hline
    \multicolumn{23}{c}{\textbf{Long Trajectory}} \\
    \hline
    \textcolor{red}{Kiko\_loop} & 314.68 & 0.90 
        & 22.81 & 7.24 & 98 & 26 & 0.87 
        & 20.12 & 6.39 & 59 & 153 & 5.11 
        & 2.51 & 0.79 & 100 & 17 & 0.56 
        & 26.79 & 8.51 & 99 & 8 & 0.28 \\ 
    \textcolor{orange}{Shopping\_loop} & 151.91 & 0.97 
        & 3.02 & 1.98 & 97 & 25 & 0.85 
        & 21.41 & 14.09 & 89 & 138 & 4.63 
        & 2.52 & 1.65 & 100 & 16 & 0.55 
        & 12.21 & 8.03 & 99 & 9 & 0.31 \\ 
    \textcolor{blue}{UZH\_HG} & 142.12 & 0.75 
        & 30.60 & 21.53 & 94 & 26 & 0.88 
        & 16.11 & 11.33 & 47 & 170 & 5.70 
        & 2.86 & 2.01 & 100 & 17 & 0.58 
        & 11.45 & 8.05 & 99 & 9 & 0.32 \\ 
    TH\_loop & 154.74 & 0.66 
        & x & x & 88 & 22 & 0.72 
        & 0.01 & 0.006 & 0.9 & 131 & 4.37 
        & 4.53 & 2.92 & 100 & 18 & 0.62 
        & 6.71 & 4.33 & 99 & 9 & 0.33 \\ 
    \hline
    \multicolumn{23}{c}{\textbf{Changing Elevation (stairs) - Short Trajectories}} \\
    \hline
    \textcolor{red}{UZH\_stairs} & 13.16 & 0.32 
        & 0.06 & 0.45 & 77 & 23 & 0.78 
        & 1.52 & 11.55 & 37 & 138 & 4.62 
        & 0.04 & 0.30 & 100 & 22 & 0.78 
        & 0.05 & 0.37 & 97 & 15 & 0.5 \\ 
    \textcolor{red}{G6\_exit} & 60.21 & 0.69 
        & 6.67 & 11.07 & 97 & 26 & 0.86 
        & x & x & x & 153 & 5.11 
        & 0.41 & 0.68 & 100 & 17 & 0.57 
        & 1.91 & 3.17 & 97 & 12 & 0.40 \\ 
    \textcolor{orange}{Museum\_up} & 12.20 & 0.36 
        & 0.42 & 3.44 & 82 & 22 & 0.75
        & 0.03 & 0.24 & 6.2 & 164 & 5.44 
        & 0.03 & 0.24 & 100 & 18 & 0.61 
        & 0.02 & 0.16 & 98 & 12 & 0.41 \\ 
    TS\_stairs & 27.67 & 0.53 
        & 0.26 & 0.93 & 94 & 25 & 0.84
        & 2.43 & 8.78 & 41 & 158 & 5.30 
        & 0.21 & 0.75 & 100 & 16 & 0.56 
        & 0.41 & 1.48 & 96 & 12 & 0.39 \\ 
    \hline
    \multicolumn{23}{c}{\textbf{Motion Transition}} \\
    \hline
    \textcolor{red}{Getoff\_pharmacy} & 159.50 & 0.86 
        & 81.01 & 50.78 & 85 & 25 & 0.83
        & 34.91 & 21.88 & 48 & 153 & 5.10 
        & 27.95 & 17.52 & 100 & 17 & 0.59
        & 27.55 & 17.27 & 98 & 10 & 0.33 \\ 
    \textcolor{red}{Ground\_53} & 32.40 & 0.53 
        & x & x & 96 & 25 & 0.84
        & 4.54 & 14.01 & 79 & 136 & 4.54 
        & 3.30 & 10.18 & 100 & 19 & 0.64 
        & 3.84 & 11.85 & 98 & 13 & 0.45 \\ 
    \textcolor{orange}{Ramp\_checkin2} & 191.77 & 0.87 
        & 7.31 & 3.81 & 98 & 26 & 0.87
        & 46.95 & 24.48 & 97 & 142 & 4.74 
        & 1.82 & 0.94 & 100 & 20 & 0.67 
        & 0.77 & 0.40 & 99 & 10 & 0.34 \\ 
    \bottomrule
	\end{tabularx}
	\begin{adjustwidth}{+\extralength}{0cm}
	\end{adjustwidth}
\end{table}
\finishlandscape
\subsection{Evaluation Metrics}
\label{sec:evaluation_metrics}
To evaluate VO and SLAM systems, we used the root mean square error (RMSE) of absolute trajectory error (ATE) \cite{sturm2012benchmark}. This metric quantifies the accuracy of the estimated trajectory compared with the ground truth by measuring the root mean square of the differences between the estimated and true poses. For ORB-SLAM3 and SVO, the absolute trajectory error was calculated using the TUM RGB-D benchmark tool \cite{TUM_evaluation_tool} and RPG trajectory evaluation tool \cite{zhang2018tutorial}, respectively. ORB-SLAM3 outputs trajectories in TUM format, which is directly compatible with the TUM RGB-D benchmark tool, whereas SVO includes the RPG evaluation tool in its package, making it the most suitable choice for evaluating SVO's output. For DROID-SLAM and DPV-SLAM, custom scripts were developed to compute the same metrics using the EVO package \cite{grupp2017evo}, a flexible tool capable of handling their output formats.
%
%

Trajectory accuracy alone does not fully capture the performance of a SLAM system, particularly in challenging real-world scenarios. SLAM systems are susceptible to initialization failures and tracking loss, particularly under conditions such as motion blur, lack of visual features, or occlusions. Such disruptions lead to gaps in the estimated trajectory, thus affecting the overall reliability of the system. To address this, we introduced the pose estimation coverage (PEC) metric, calculated as (number of estimated poses/total number of frames) × 100. The PEC quantifies the system's ability to maintain consistent pose estimation throughout a sequence, offering insights into its robustness against initialization failures and tracking losses. A notably low PEC often indicates critical failure, which can make the ATE unreliable.
%
%
To further evaluate each system, we introduced a drift percentage (DP) metric, defined as (ATE/trajectory length) × 100. This metric quantifies the drift of the system as a percentage of the total distance traveled. A lower value indicates better performance, with the system maintaining more accurate localization over extended trajectories.
%
%
To ensure a comprehensive evaluation that considers both accuracy and robustness, we adopted a dual-criteria approach that incorporates both ATE and PEC. A sequence is considered successful if the drift percentage (DP) value is less than 1\% of the distance traveled and its PEC exceeds 90\%. 

%
%
Beyond accuracy and robustness, real-time performance is crucial for visually impaired navigation applications because it directly influences user experience and safety. To directly relate system performance to the end-user experience, we developed a metric that compares the processing speed of SLAM systems with the user’s walking speed. The fundamental idea is that, if the system can process more distance than the user walks in a given time interval (e.g., one second), it is considered to be performing in real time. First, we calculated the frames per second (FPS) processed by the system, indicating the number of frames handled per second. Next, we calculated the distance per frame (DPF), which represents the distance between consecutive camera frames. DPF was calculated using the formula DPF = AWS /camera\_fps, where AWS is the average user walking speed and camera\_fps is the camera's frame rate. Finally, we introduced the processed distance rate (PDR), which is obtained by multiplying the FPS and DPF. If the PDR meets or exceeds the average walking speed, the system is considered capable of keeping pace with the user's movement, indicating adequate real-time performance for BVI navigation. We quantified this capability through the real-time factor (RTF), defined as the ratio of PDR to AWS, where the values $\geq 1$ demonstrate the real-time processing capacity.
%
%
\subsection{Evaluation Results}
\label{sec:evaluation_results}
The evaluation results presented in Table~\ref{tab:evaluation_results} demonstrate that InCrowd-VI successfully captures challenging scenarios that push the limits of the current state-of-the-art VO and SLAM systems. Our analysis of these results is two-fold. First, we assessed the system performance against the key requirements of visually impaired navigation applications. Second, we analyzed the impact of specific environmental factors, such as crowd density, trajectory length, lighting conditions, and motion transition, on system performance. This analysis validates the effectiveness of InCrowd-VI as a benchmark, while also identifying critical areas that require further research to make SLAM viable for visually impaired navigation.
 
To effectively address the needs of BVI individuals during navigation, several key requirements must be satisfied, including real-time performance, high localization accuracy, and robustness. In addition, the system should maintain long-distance consistency with minimal drifts.
For this study, we established a localization accuracy criterion of 0.5 meters for indoor environments, which we consider suitable for BVI navigation applications. Across the sequences, ORB-SLAM3 and DROID-SLAM generally met or approached this criterion in low-density crowds. However, under more challenging conditions, both classical and learning-based systems exhibit a higher ATE, which significantly surpasses the required accuracy. 
Although not directly comparable to the requirement for robustness to occlusions, dynamic objects, and other challenging conditions, pose estimation coverage (PEC) provides insights into system reliability. Robust pose estimation is essential for maintaining consistent localization. BVI navigation requires minimal interruption during the operation. Deep learning-based approaches consistently achieved high PEC values ($>90\%$) across all sequences, whereas classical approaches failed to maintain consistent tracking, particularly in scenes with lighting variations. This indicates a limitation in handling the dynamics in the environment, which is critical for real-world BVI applications.

Real-time performance requires the system to match or exceed the walking speed of the user. Our processed distance rate (PDR) metric, when compared to the average walking speed (AWS), indicates that classical approaches generally achieve real-time or near real-time performance across almost all sequences. However, deep learning-based methods fall short of real-time processing, indicating delays that could hinder real-time feedback for users. 
Regarding long-distance consistency, the drift percentage (DP) metric revealed that both classical and deep learning-based approaches frequently exceeded the desired 1\% drift threshold, particularly for longer trajectories. Classical approaches typically show higher drift percentages (often exceeding 5-10\% of the trajectory length), whereas deep learning-based methods generally maintain lower drift rates but still struggle to consistently achieve the desired threshold.
It is important to note that, although some systems may approach or meet certain requirements under controlled conditions, they all face significant challenges in maintaining consistent performance across diverse real-world scenarios represented in the InCrowd-VI dataset. This highlights the gap between the current VO and SLAM capabilities and requirements for visually impaired navigation in crowded indoor environments.

Beyond these requirements, our analysis reveals how specific environmental factors affect the system performance.
The evaluation results highlight the significant impact of crowd density on the performance of VO and SLAM systems. In scenarios with high crowd density, systems often exhibit a higher ATE than sequences with low or no crowd density. The trajectory length also plays a crucial role. For instance, in the Entrance\_checkin1 scene, despite the medium crowd density, the system reported a lower error owing to the short trajectory length. In addition to crowd density, other challenging conditions also affect system performance. The TS116 scene, which is a typical university building environment with artificial lighting and sunlight in some areas, exemplifies the impact of lighting variations. Despite the absence of pedestrians, ORB-SLAM3 encountered difficulties during initialization and experienced frequent tracking losses owing to lighting challenges. Similarly, in the TH\_loop scene, the user's proximity to a wall causing occlusions, combined with challenging lighting conditions from large windows, caused tracking issues, even without the presence of pedestrians.

\begin{figure}[H]
    \centering
        \begin{subfigure}{0.45\textwidth}
            \includegraphics[width=\textwidth]{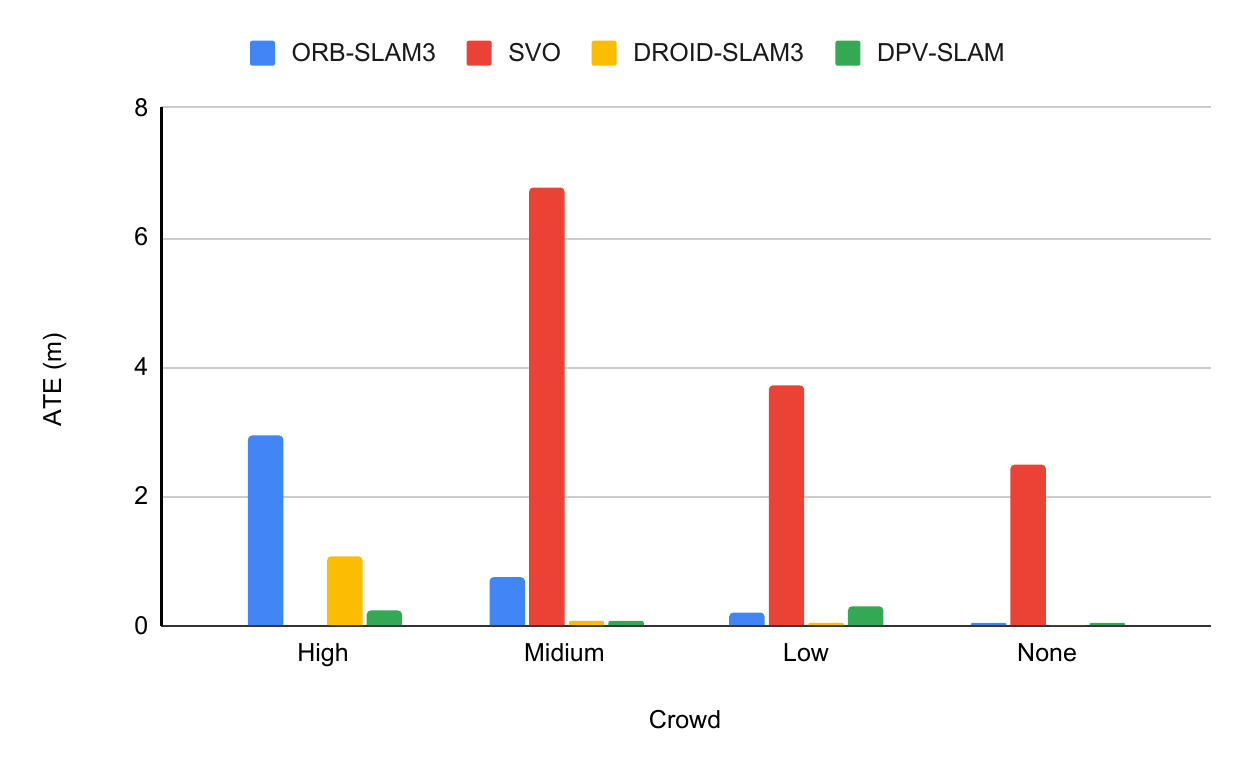}
            \caption{Crowd}
            \label{fig:crowd}
        \end{subfigure}
        \hfill
        \begin{subfigure}{0.45\textwidth}
            \includegraphics[width=\textwidth]{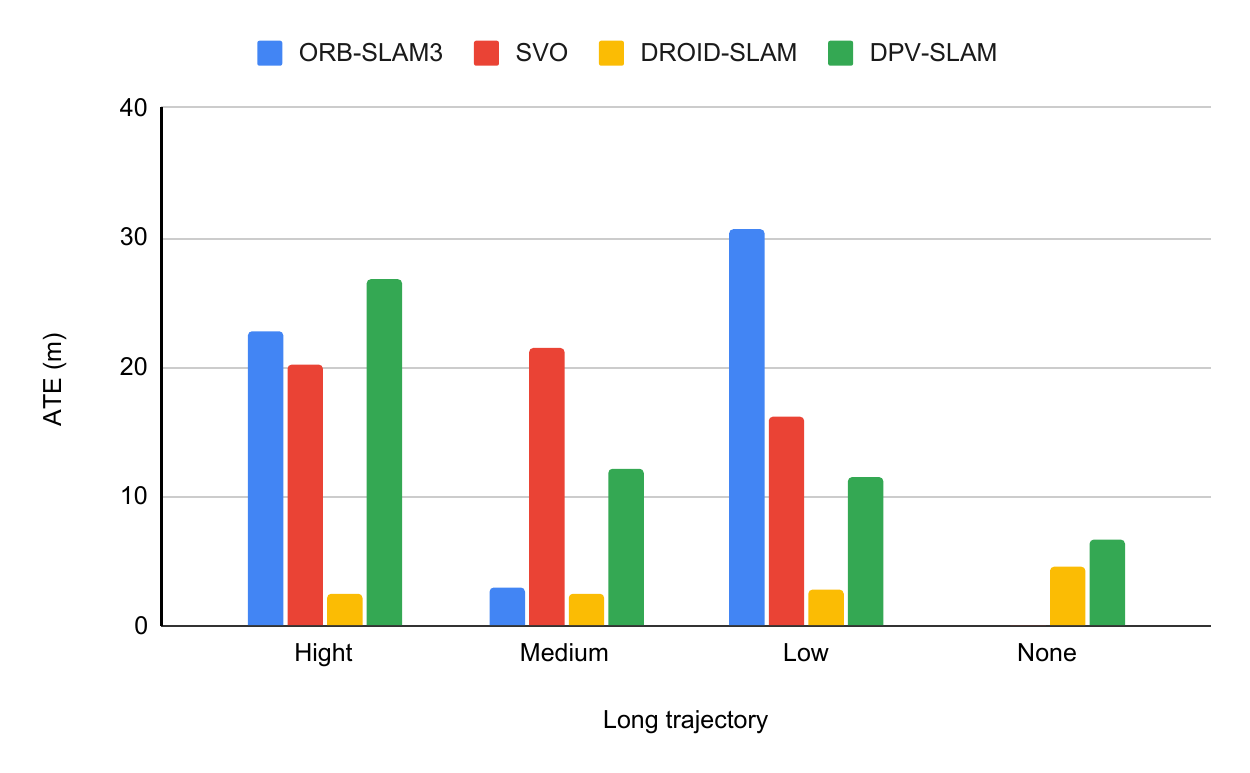}
            \caption{Long Trajectory}
            \label{fig:long_trajectory}
        \end{subfigure}
        \begin{subfigure}{0.45\textwidth}
            \includegraphics[width=\textwidth]{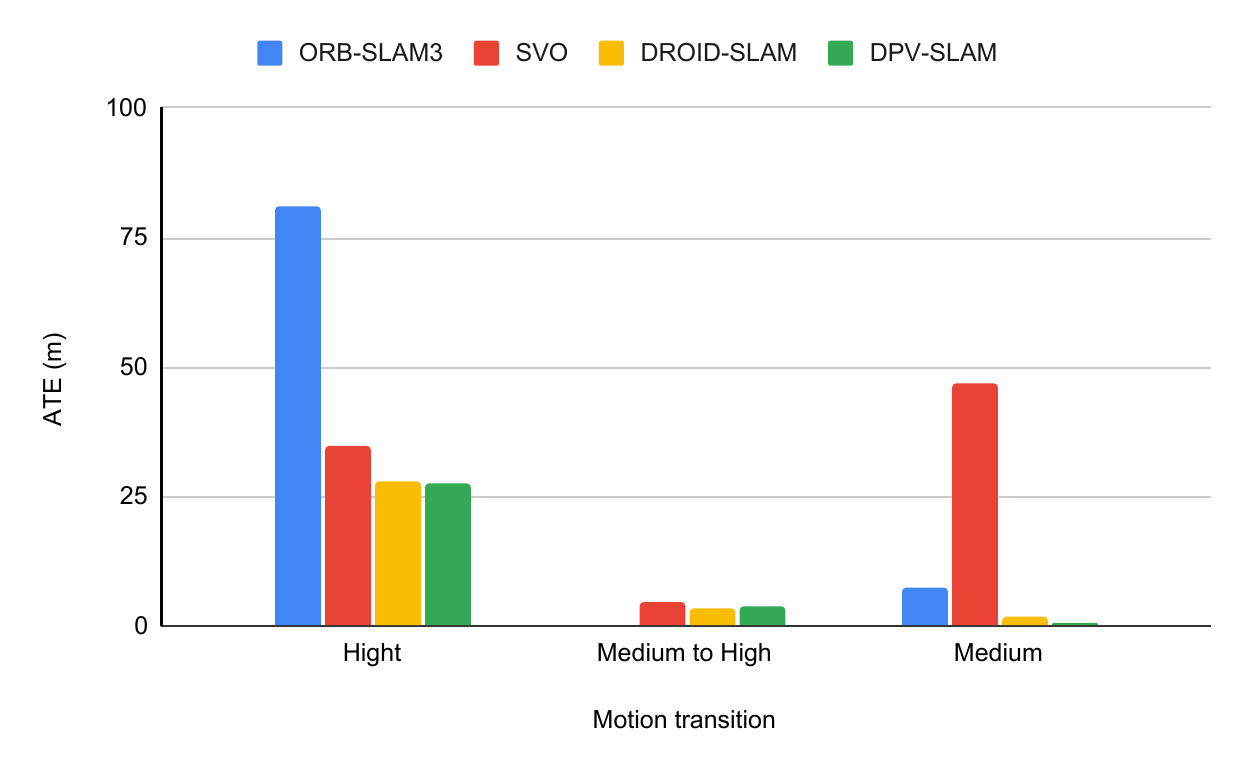}
            \caption{Motion Transition}
            \label{fig:motion_transition}
        \end{subfigure}
        \hfill
        \begin{subfigure}{0.45\textwidth}
            \includegraphics[width=\textwidth]{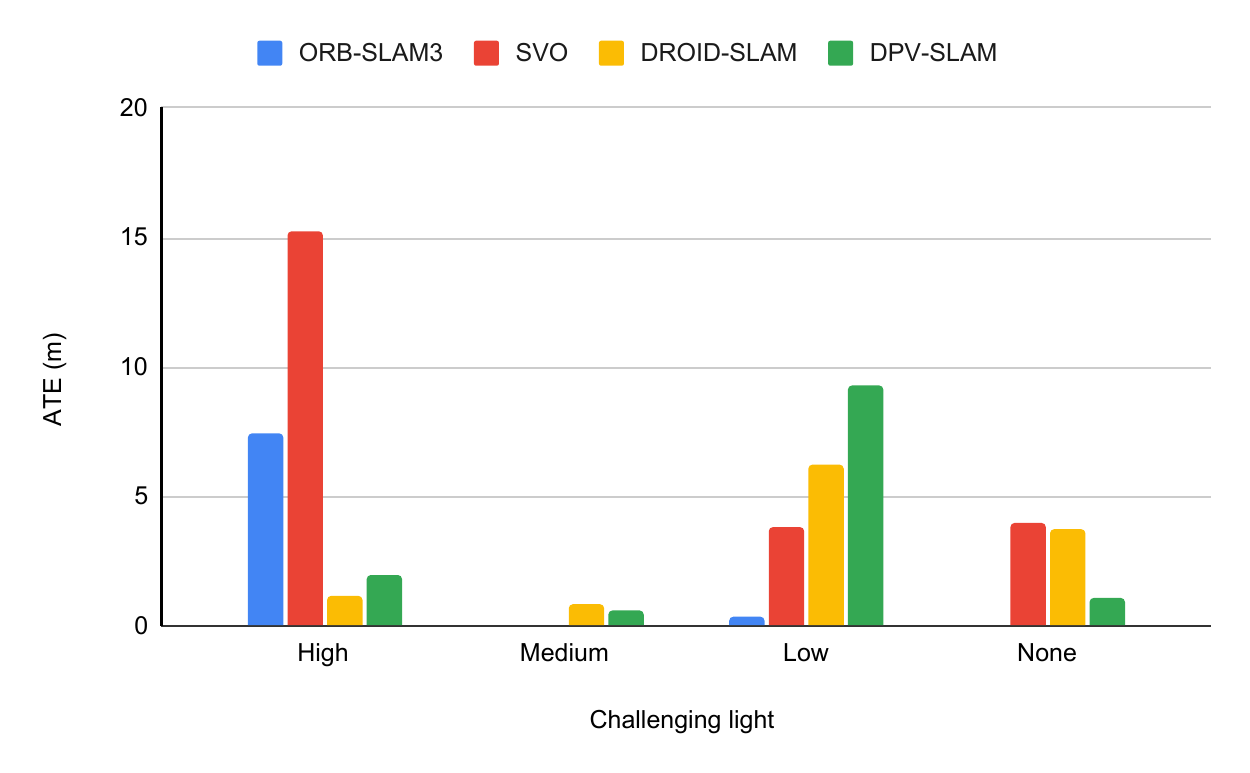}
            \caption{Challenging Light}
            \label{fig:challenging_light}
        \end{subfigure}
        \begin{subfigure}{0.45\textwidth}
            \includegraphics[width=\textwidth]{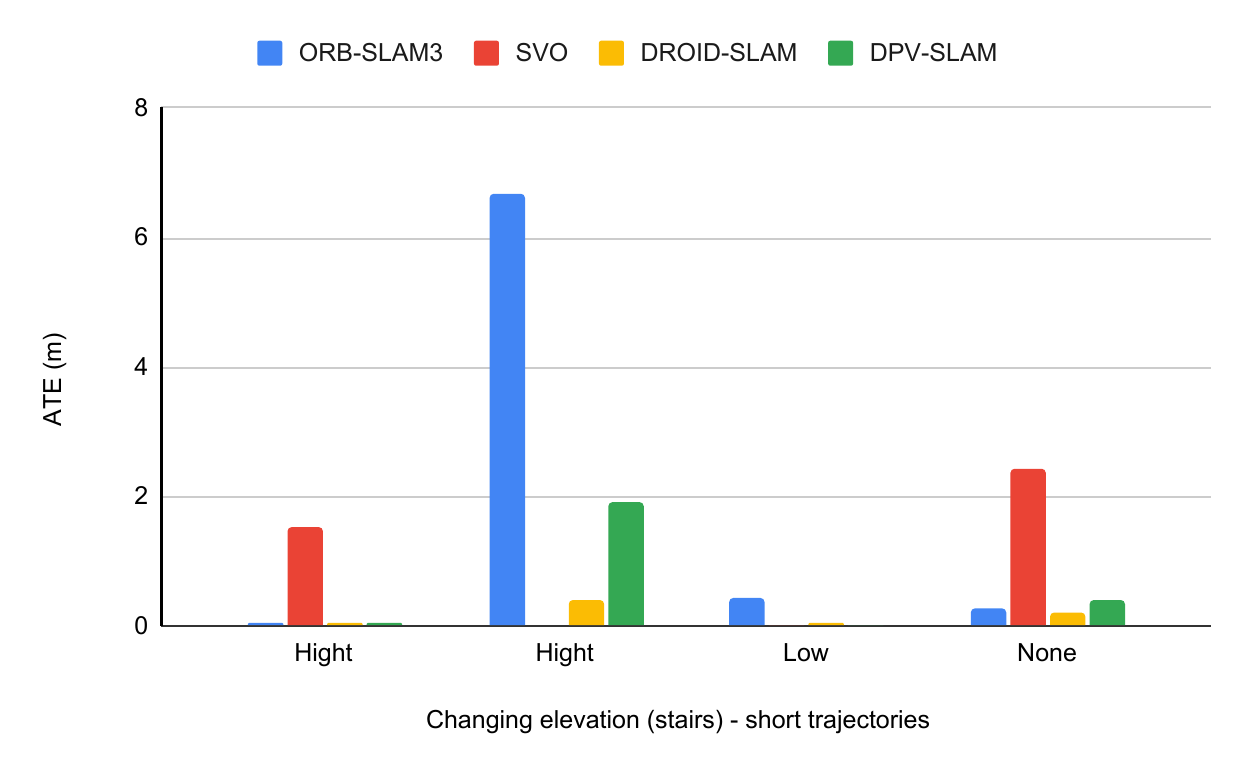}
            \caption{Changing Elevation (Stairs)}
            \label{fig:stairs}
        \end{subfigure}
        \hfill
        \begin{subfigure}{0.45\textwidth}
            \includegraphics[width=\textwidth]{images/emty.png}
        \end{subfigure}
    \caption{ATE comparison of evaluated SLAM systems under challenging conditions, with the x-axis depicting sequences categorized by crowd density: high, medium, low, and none.\label{fig:ATE_charts}}
\end{figure}

The results also demonstrated the influence of specific scenario characteristics. In scenes involving stairs, the short trajectory lengths allow the systems to handle elevation changes effectively, despite the inherent challenges. However, in scenes with motion transitions, particularly Getoff\_pharmacy, where the user transitions from a moving train to a stationary platform, the systems struggle to differentiate between ego-motion and environmental motion, resulting in a poor performance. Figure~\ref{fig:ATE_charts} provides a visual representation of the impact of these challenging conditions on ATE across different systems. The charts reveal that deep learning-based methods generally maintain a lower ATE across most challenging conditions, whereas classical approaches tend to show more dramatic accuracy degradation with increasing environmental complexity.
In summary, our evaluation revealed three key findings: (1) deep-learning-based methods demonstrate superior robustness but struggle with real-time processing, whereas classical approaches offer better processing speed but lack consistency in challenging conditions; (2) environmental factors, such as crowd density and lighting variations, significantly impact all systems, with performance degradation increasing in proportion to crowd density; and (3) none of the evaluated systems fully satisfied the combined requirements for reliable BVI navigation in terms of accuracy, robustness, and real-time performance in complex indoor environments. These findings underscore the need for new approaches that can better balance these competing demands while maintaining reliability across diverse real-world conditions. 
\section{Conclusion}
\label{sec:conclusion}
This paper introduces InCrowd-VI, a novel visual-inertial dataset designed to address the challenges of SLAM in indoor pedestrian-rich environments, particularly for visually impaired navigation. Our evaluation of the state-of-the-art VO and SLAM algorithms on InCrowd-VI demonstrated significant performance limitations across both classical and deep learning approaches,validating that the dataset effectively captures challenging real-world scenarios. These systems struggle with the complexities of crowded scenes, lighting variations, and motion transitions, highlighting the gap between the current capabilities and real-world indoor navigation demands. While this study highlights significant performance limitations of SLAM and VO algorithms on the InCrowd-VI dataset, a more detailed analysis of specific failure modes (e.g., feature loss, motion blur, and occlusion) remains an important direction for future work. Such an analysis could provide researchers with actionable insights to develop algorithms that are better suited to challenging indoor navigation scenarios.

InCrowd-VI serves as a crucial benchmark for advancing SLAM research in complex, crowded indoor settings. It provides realistic, user-centric data that closely mirrors the challenges faced by visually impaired individuals navigating such environments. Future studies should focus on addressing the key challenges identified by the InCrowd-VI dataset. 
Although InCrowd-VI is a valuable resource for indoor SLAM research, it is important to acknowledge its limitations. The absence of depth information restricts its applicability for testing depth-based SLAM systems, and its focus on indoor environments limits its utility in outdoor and mixed-environment scenarios.
\vspace{6pt} 




\authorcontributions{Conceptualization, M.B.; methodology, M.B.; software, M.B.; validation, M.B.; formal analysis, M.B.; investigation, M.B.; resources, M.B.; data curation, M.B.; writing—original draft preparation, M.B.; writing—review and editing, M.B., H.-P.H.; visualization, M.B.; supervision, H.-P.H. and A.D.; funding acquisition, A.D. All authors have read and agreed to the published version of the manuscript.}

\funding{This research was funded by SNSF (Swiss National Science Foundation) grant number 40B2-0\_194677. The APC was funded by ZHAW Open Acess, Zurich University of Applied Sciences.}

\institutionalreview{Not applicable}

\informedconsent{Not applicable}

\dataavailability{The InCrowd-VI dataset and associated tools are publicly available at https://incrowd-vi.cloudlab.zhaw.ch/. The repository includes the complete dataset, custom tools for data extraction and processing from Meta .vrs recording files, and documentation for dataset usage.} 




\acknowledgments{The authors would like to thank the Robotics and Perception Group at the Department of Informatics, University of Zurich, for providing the Meta Aria glasses used in this study. The authors also extend their gratitude to Giovanni Cioffi for his valuable feedback.}

\conflictsofinterest{The authors declare no conflicts of interest.} 

\begin{adjustwidth}{-\extralength}{0cm}

\reftitle{References}


\bibliography{main}

\begin{thebibliography}{999}

\bibitem[Durrant-Whyte and Bailey(2006)]{durrant2006simultaneous}
Durrant-Whyte, H.; Bailey, T.
\newblock Simultaneous localization and mapping: part I.
\newblock {\em IEEE robotics \& automation magazine} {\bf 2006}, {\em 13},~99--110.

\bibitem[Cadena et~al.(2016)Cadena, Carlone, Carrillo, Latif, Scaramuzza, Neira, Reid, and Leonard]{cadena2016past}
Cadena, C.; Carlone, L.; Carrillo, H.; Latif, Y.; Scaramuzza, D.; Neira, J.; Reid, I.; Leonard, J.J.
\newblock Past, present, and future of simultaneous localization and mapping: Toward the robust-perception age.
\newblock {\em IEEE Transactions on robotics} {\bf 2016}, {\em 32},~1309--1332.

\bibitem[Pellerito et~al.(2024)Pellerito, Cannici, Gehrig, Belhadj, Dubois-Matra, Casasco, and Scaramuzza]{pelleritodeep}
Pellerito, R.; Cannici, M.; Gehrig, D.; Belhadj, J.; Dubois-Matra, O.; Casasco, M.; Scaramuzza, D.
\newblock Deep Visual Odometry with Events and Frames.
\newblock In Proceedings of the IEEE/RSJ International Conference on Intelligent Robots and Systems (IROS. IEEE,  2024.

\bibitem[Bresson et~al.(2017)Bresson, Alsayed, Yu, and Glaser]{bresson2017simultaneous}
Bresson, G.; Alsayed, Z.; Yu, L.; Glaser, S.
\newblock Simultaneous localization and mapping: A survey of current trends in autonomous driving.
\newblock {\em IEEE Transactions on Intelligent Vehicles} {\bf 2017}, {\em 2},~194--220.

\bibitem[Hanover et~al.(2024)Hanover, Loquercio, Bauersfeld, Romero, Penicka, Song, Cioffi, Kaufmann, and Scaramuzza]{hanover2024autonomous}
Hanover, D.; Loquercio, A.; Bauersfeld, L.; Romero, A.; Penicka, R.; Song, Y.; Cioffi, G.; Kaufmann, E.; Scaramuzza, D.
\newblock Autonomous drone racing: A survey.
\newblock {\em IEEE Transactions on Robotics} {\bf 2024}.

\bibitem[Alberico et~al.(2024)Alberico, Delaune, Cioffi, and Scaramuzza]{alberico2024structure}
Alberico, I.; Delaune, J.; Cioffi, G.; Scaramuzza, D.
\newblock Structure-Invariant Range-Visual-Inertial Odometry.
\newblock {\em arXiv preprint arXiv:2409.04633} {\bf 2024}.

\bibitem[Somasundaram et~al.(2023)Somasundaram, Dong, Tang, Straub, Yan, Goesele, Engel, De~Nardi, and Newcombe]{somasundaram2023project}
Somasundaram, K.; Dong, J.; Tang, H.; Straub, J.; Yan, M.; Goesele, M.; Engel, J.J.; De~Nardi, R.; Newcombe, R.
\newblock Project Aria: A new tool for egocentric multi-modal AI research.
\newblock {\em arXiv preprint arXiv:2308.13561} {\bf 2023}.

\bibitem[Zhang et~al.(2022)Zhang, Helmberger, Fu, Wisth, Camurri, Scaramuzza, and Fallon]{zhang2022hilti}
Zhang, L.; Helmberger, M.; Fu, L.F.T.; Wisth, D.; Camurri, M.; Scaramuzza, D.; Fallon, M.
\newblock Hilti-oxford dataset: A millimeter-accurate benchmark for simultaneous localization and mapping.
\newblock {\em IEEE Robotics and Automation Letters} {\bf 2022}, {\em 8},~408--415.

\bibitem[Geiger et~al.(2012)Geiger, Lenz, and Urtasun]{geiger2012we}
Geiger, A.; Lenz, P.; Urtasun, R.
\newblock Are we ready for autonomous driving? the kitti vision benchmark suite.
\newblock In Proceedings of the 2012 IEEE conference on computer vision and pattern recognition. IEEE,  2012, pp. 3354--3361.

\bibitem[Burri et~al.(2016)Burri, Nikolic, Gohl, Schneider, Rehder, Omari, Achtelik, and Siegwart]{burri2016euroc}
Burri, M.; Nikolic, J.; Gohl, P.; Schneider, T.; Rehder, J.; Omari, S.; Achtelik, M.W.; Siegwart, R.
\newblock The EuRoC micro aerial vehicle datasets.
\newblock {\em The International Journal of Robotics Research} {\bf 2016}, {\em 35},~1157--1163.

\bibitem[Pfrommer et~al.(2017)Pfrommer, Sanket, Daniilidis, and Cleveland]{pfrommer2017penncosyvio}
Pfrommer, B.; Sanket, N.; Daniilidis, K.; Cleveland, J.
\newblock Penncosyvio: A challenging visual inertial odometry benchmark.
\newblock In Proceedings of the 2017 IEEE International Conference on Robotics and Automation (ICRA). IEEE,  2017, pp. 3847--3854.

\bibitem[Majdik et~al.(2017)Majdik, Till, and Scaramuzza]{majdik2017zurich}
Majdik, A.L.; Till, C.; Scaramuzza, D.
\newblock The Zurich urban micro aerial vehicle dataset.
\newblock {\em The International Journal of Robotics Research} {\bf 2017}, {\em 36},~269--273.

\bibitem[Li et~al.(2018)Li, Saeedi, McCormac, Clark, Tzoumanikas, Ye, Huang, Tang, and Leutenegger]{li2018interiornet}
Li, W.; Saeedi, S.; McCormac, J.; Clark, R.; Tzoumanikas, D.; Ye, Q.; Huang, Y.; Tang, R.; Leutenegger, S.
\newblock Interiornet: Mega-scale multi-sensor photo-realistic indoor scenes dataset.
\newblock {\em arXiv preprint arXiv:1809.00716} {\bf 2018}.

\bibitem[Schubert et~al.(2018)Schubert, Goll, Demmel, Usenko, St{\"u}ckler, and Cremers]{schubert2018tum}
Schubert, D.; Goll, T.; Demmel, N.; Usenko, V.; St{\"u}ckler, J.; Cremers, D.
\newblock The TUM VI benchmark for evaluating visual-inertial odometry.
\newblock In Proceedings of the 2018 IEEE/RSJ International Conference on Intelligent Robots and Systems (IROS). IEEE,  2018, pp. 1680--1687.

\bibitem[Delmerico et~al.(2019)Delmerico, Cieslewski, Rebecq, Faessler, and Scaramuzza]{delmerico2019we}
Delmerico, J.; Cieslewski, T.; Rebecq, H.; Faessler, M.; Scaramuzza, D.
\newblock Are we ready for autonomous drone racing? the UZH-FPV drone racing dataset.
\newblock In Proceedings of the 2019 International Conference on Robotics and Automation (ICRA). IEEE,  2019, pp. 6713--6719.

\bibitem[Ramezani et~al.(2020)Ramezani, Wang, Camurri, Wisth, Mattamala, and Fallon]{ramezani2020newer}
Ramezani, M.; Wang, Y.; Camurri, M.; Wisth, D.; Mattamala, M.; Fallon, M.
\newblock The newer college dataset: Handheld lidar, inertial and vision with ground truth.
\newblock In Proceedings of the 2020 IEEE/RSJ International Conference on Intelligent Robots and Systems (IROS). IEEE,  2020, pp. 4353--4360.

\bibitem[Minoda et~al.(2021)Minoda, Schilling, W{\"u}est, Floreano, and Yairi]{minoda2021viode}
Minoda, K.; Schilling, F.; W{\"u}est, V.; Floreano, D.; Yairi, T.
\newblock Viode: A simulated dataset to address the challenges of visual-inertial odometry in dynamic environments.
\newblock {\em IEEE Robotics and Automation Letters} {\bf 2021}, {\em 6},~1343--1350.

\bibitem[Lee et~al.(2021)Lee, Ryu, Yeon, Lee, Kim, Han, Cabon, Weinzaepfel, Gu{\'e}rin, Csurka, et~al.]{lee2021large}
Lee, D.; Ryu, S.; Yeon, S.; Lee, Y.; Kim, D.; Han, C.; Cabon, Y.; Weinzaepfel, P.; Gu{\'e}rin, N.; Csurka, G.;  et~al.
\newblock Large-scale localization datasets in crowded indoor spaces.
\newblock In Proceedings of the Proceedings of the IEEE/CVF Conference on Computer Vision and Pattern Recognition,  2021, pp. 3227--3236.

\bibitem[Bojko et~al.(2022)Bojko, Dupont, Tamaazousti, and Borgne]{bojko2022self}
Bojko, A.; Dupont, R.; Tamaazousti, M.; Borgne, H.L.
\newblock Self-improving SLAM in dynamic environments: learning when to mask.
\newblock {\em arXiv preprint arXiv:2210.08350} {\bf 2022}.

\bibitem[Zhang et~al.(2023)Zhang, An, Shi, Wang, Wei, Zhang, Meng, Sun, Wang, Liang, et~al.]{zhang2023cid}
Zhang, Y.; An, N.; Shi, C.; Wang, S.; Wei, H.; Zhang, P.; Meng, X.; Sun, Z.; Wang, J.; Liang, W.;  et~al.
\newblock CID-SIMS: Complex indoor dataset with semantic information and multi-sensor data from a ground wheeled robot viewpoint.
\newblock {\em The International Journal of Robotics Research} {\bf 2023}, p. 02783649231222507.

\bibitem[Recchiuto et~al.(2017)Recchiuto, Scalmato, and Sgorbissa]{recchiuto2017dataset}
Recchiuto, C.T.; Scalmato, A.; Sgorbissa, A.
\newblock A dataset for human localization and mapping with wearable sensors.
\newblock {\em Robotics and Autonomous Systems} {\bf 2017}, {\em 97},~136--143.

\bibitem[Cort{\'e}s et~al.(2018)Cort{\'e}s, Solin, Rahtu, and Kannala]{cortes2018advio}
Cort{\'e}s, S.; Solin, A.; Rahtu, E.; Kannala, J.
\newblock ADVIO: An authentic dataset for visual-inertial odometry.
\newblock In Proceedings of the Proceedings of the European Conference on Computer Vision (ECCV),  2018, pp. 419--434.

\bibitem[Charatan et~al.(2022)Charatan, Fan, and Kimia]{charatan2022benchmarking}
Charatan, D.; Fan, H.; Kimia, B.
\newblock Benchmarking Pedestrian Odometry: The Brown Pedestrian Odometry Dataset (BPOD).
\newblock In Proceedings of the 2022 International Conference on 3D Vision (3DV). IEEE,  2022, pp. 1--11.

\bibitem[Liu et~al.(2021)Liu, Fu, Chen, Goossens, Tao, and Zhao]{liu2021simultaneous}
Liu, Y.; Fu, Y.; Chen, F.; Goossens, B.; Tao, W.; Zhao, H.
\newblock Simultaneous localization and mapping related datasets: A comprehensive survey.
\newblock {\em arXiv preprint arXiv:2102.04036} {\bf 2021}.

\bibitem[Tosi et~al.(2024)Tosi, Zhang, Gong, Sandstr{\"o}m, Mattoccia, Oswald, and Poggi]{tosi2024nerfs}
Tosi, F.; Zhang, Y.; Gong, Z.; Sandstr{\"o}m, E.; Mattoccia, S.; Oswald, M.R.; Poggi, M.
\newblock How NeRFs and 3D Gaussian Splatting are Reshaping SLAM: a Survey.
\newblock {\em arXiv preprint arXiv:2402.13255} {\bf 2024}.

\bibitem[{Meta AI Research}(2024)]{projectariasensorspec}
{Meta AI Research}.
\newblock Project Aria Hardware Specifications: \url{https://facebookresearch.github.io/projectaria_tools/docs/tech_spec/hardware_spec},  2024.
\newblock [Online; accessed 2024-09-04].

\bibitem[{Meta AI Research}({\natexlab{a}})]{CameraIntrinsicModelsforProjectAriadevices}
{Meta AI Research}.
\newblock Camera Intrinsic Models for Project Aria devices: \url{https://facebookresearch.github.io/projectaria_tools/docs/tech_insights/camera_intrinsic_models}.
\newblock [Online; accessed 2024-09-04].

\bibitem[{Meta AI Research}({\natexlab{b}})]{metaariaimu}
{Meta AI Research}.
\newblock Camera Intrinsic Models for Project Aria devices: \url{https://facebookresearch.github.io/projectaria_tools/docs/tech_insights/sensor_measurement_model}.
\newblock [Online; accessed 2024-09-04].

\bibitem[{Meta AI Research}({\natexlab{c}})]{3DCoordinateFrameConventionsforProjectAriaGlasses}
{Meta AI Research}.
\newblock 3D Coordinate Frame Conventions for Project Aria Glasses: \url{https://facebookresearch.github.io/projectaria_tools/docs/data_formats/coordinate_convention/3d_coordinate_frame_convention}.
\newblock [Online; accessed 2024-09-04].

\bibitem[Schops et~al.(2019)Schops, Sattler, and Pollefeys]{schops2019bad}
Schops, T.; Sattler, T.; Pollefeys, M.
\newblock Bad slam: Bundle adjusted direct rgb-d slam.
\newblock In Proceedings of the Proceedings of the IEEE/CVF Conference on Computer Vision and Pattern Recognition,  2019, pp. 134--144.

\bibitem[Bennett et~al.(2019)Bennett, Valenzuela, Fleenor, Morrison, and Haegele]{bennett2019walking}
Bennett, H.J.; Valenzuela, K.A.; Fleenor, K.; Morrison, S.; Haegele, J.A.
\newblock Walking biomechanics and energetics of individuals with a visual impairment: a preliminary report.
\newblock {\em Human Movement} {\bf 2019}, {\em 20},~8--18.

\bibitem[Teed and Deng(2021)]{teed2021droid}
Teed, Z.; Deng, J.
\newblock Droid-slam: Deep visual slam for monocular, stereo, and rgb-d cameras.
\newblock {\em Advances in neural information processing systems} {\bf 2021}, {\em 34},~16558--16569.

\bibitem[Lipson et~al.(2024)Lipson, Teed, and Deng]{lipson2024deep}
Lipson, L.; Teed, Z.; Deng, J.
\newblock Deep Patch Visual SLAM.
\newblock {\em arXiv preprint arXiv:2408.01654} {\bf 2024}.

\bibitem[Campos et~al.(2021)Campos, Elvira, Rodr{\'\i}guez, Montiel, and Tard{\'o}s]{campos2021orb}
Campos, C.; Elvira, R.; Rodr{\'\i}guez, J.J.G.; Montiel, J.M.; Tard{\'o}s, J.D.
\newblock Orb-slam3: An accurate open-source library for visual, visual--inertial, and multimap slam.
\newblock {\em IEEE Transactions on Robotics} {\bf 2021}, {\em 37},~1874--1890.

\bibitem[Forster et~al.(2016)Forster, Zhang, Gassner, Werlberger, and Scaramuzza]{forster2016svo}
Forster, C.; Zhang, Z.; Gassner, M.; Werlberger, M.; Scaramuzza, D.
\newblock SVO: Semidirect visual odometry for monocular and multicamera systems.
\newblock {\em IEEE Transactions on Robotics} {\bf 2016}, {\em 33},~249--265.

\bibitem[Teed et~al.(2024)Teed, Lipson, and Deng]{teed2024deep}
Teed, Z.; Lipson, L.; Deng, J.
\newblock Deep patch visual odometry.
\newblock {\em Advances in Neural Information Processing Systems} {\bf 2024}, {\em 36}.

\bibitem[Sturm et~al.(2012)Sturm, Engelhard, Endres, Burgard, and Cremers]{sturm2012benchmark}
Sturm, J.; Engelhard, N.; Endres, F.; Burgard, W.; Cremers, D.
\newblock A benchmark for the evaluation of RGB-D SLAM systems.
\newblock In Proceedings of the 2012 IEEE/RSJ international conference on intelligent robots and systems. IEEE,  2012, pp. 573--580.

\bibitem[Computer Vision Group TUM School~of Computation and of~Munich()]{TUM_evaluation_tool}
Computer Vision Group TUM School~of Computation, I.; of~Munich, T.T.U.
\newblock Useful tools for the RGB-D benchmark.
\newblock \url{https://edit.paperpal.com/documents/a108dfc5-1231-4f54-a402-15d7c7b5f556}.

\bibitem[Zhang and Scaramuzza(2018)]{zhang2018tutorial}
Zhang, Z.; Scaramuzza, D.
\newblock A tutorial on quantitative trajectory evaluation for visual (-inertial) odometry.
\newblock In Proceedings of the 2018 IEEE/RSJ International Conference on Intelligent Robots and Systems (IROS). IEEE,  2018, pp. 7244--7251.

\bibitem[Grupp(2017)]{grupp2017evo}
Grupp, M.
\newblock evo: Python package for the evaluation of odometry and SLAM.
\newblock \url{https://github.com/MichaelGrupp/evo},  2017.

\end{thebibliography}

\PublishersNote{}
\end{adjustwidth}
\end{document}